%% file: MAIN_compile_this_one.tex
\documentclass[conference]{IEEEtran}

\input{definitions}

\usepackage{bm}
\usepackage{cleveref}

\usepackage{amsfonts}
\usepackage{amssymb}
\usepackage{amsbsy}
\usepackage{amsmath}

\IEEEoverridecommandlockouts                              % This command is only needed if 
\title{Convex Geometric Motion Planning on Lie Groups via Moment Relaxation}

\author{Sangli Teng, Ashkan Jasour, Ram Vasudevan, Maani Ghaffari
\thanks{S.~Teng, R.~Vasudevan, and M.~Ghaffari are with the University of Michigan, Ann Arbor, MI 48109, USA. {\tt\small\{sanglit,ramv,maanigj\} @umich.edu}}
\thanks{A.~Jasour is with Team 347T-Robotic Aerial Mobility, Jet Propulsion Lab, Pasadena, CA, 91109. {\tt\small jasour@jpl.caltech.edu}}%
}

\begin{document}

\maketitle

\begin{abstract}
% Selling point: Combinatorial problems can hardly deal with configuration states. Fixed runtime. Certifiable optimality. Though global optimization techniques, such as combinatorial methods, is complete, the runtime is exponential with respect to the number of variables and dimensions. However, the configuration space of the rigid body can only be approximated by binary intervals at the expense of increasing variables. 
This paper reports a novel result: with proper robot models on matrix Lie groups, one can formulate the kinodynamic motion planning problem for rigid body systems as \emph{exact} polynomial optimization problems that can be relaxed as semidefinite programming (SDP). Due to the nonlinear rigid body dynamics, the motion planning problem for rigid body systems is nonconvex. Existing global optimization-based methods do not properly deal with the configuration space of the 3D rigid body; thus, they do not scale well to long-horizon planning problems. We use Lie groups as the configuration space in our formulation and apply the variational integrator to formulate the forced rigid body systems as quadratic polynomials. Then we leverage Lasserre's hierarchy to obtain the globally optimal solution via SDP. By constructing the motion planning problem in a sparse manner, the results show that the proposed algorithm has \emph{linear} complexity with respect to the planning horizon. This paper demonstrates the proposed method can provide rank-one optimal solutions at relaxation order two for most of the testing cases of 1) 3D drone landing using the full dynamics model and 2) inverse kinematics for serial manipulators. % \tsl{I softened the tightness claim.} % The results show that the proposed moment relaxation is tight at order two  % The result is hopeful that one could solve general % For rigid body dynamics, converting the configuration space to binary intervals will not be computationally tractable
% To solve this problem, we formulate that fully exploits the geometric space of robotics systems. We show that the proposed formulations are quadratic polynomials for forced rigid body systems. using variational integrators
\end{abstract} % Kinodynamics motion planning that aims to synthesize robot motion subject to nonlinear dynamics constraints is fundamental in robotics research

\IEEEpeerreviewmaketitle
\section{Introduction}

% \tsl{Why global solution is needed?}

% \tsl{Why we want globally optimal dynamic model?}

% \tsl{Why moment relaxation method is good? }

% \tsl{First dynamic level. }
% This paper formulates the kinodynamic motion planning problem as general moment problem (GMP)
The kinodynamic motion planning \cite{donald1993kinodynamic}, or trajectory optimization \cite{betts1998survey}, which aims to synthesize robot motion subject to kinematics, dynamics, and input constraints, is fundamental in robotics research. A typical formulation of kinodynamic motion planning is a constrained optimization problem, usually nonconvex due to the nonlinear dynamics and obstacle configurations. Despite the nonconvexity of these problems, the optimization methods that exploit the local gradient information have been successfully applied to find local solutions. Unless problem-specific convexification is accessible~\cite{accikmecse2013lossless}, there is generally no guarantee on the global optimality %. As these methods are only based on gradients, there is no guarantee of global optimality.

Indeed, the complexity of motion planning problems with arbitrary obstacles is high \cite{reif1979complexity, hopcroft1984complexity, canny1987new} that one should not expect an efficient algorithm for general problems. For motion planning problems in moderate size, global optimization techniques, such as mixed integer programming \cite{richards2002aircraft, schouwenaars2001mixed, deits2015efficient, ding2020kinodynamic, ding2018single, deits2014footstep, dai2019global} and polynomial optimization \cite{el2021piecewise, amice2023finding, trutman2022globally}, have been applied to obtain or approximate the globally optimal solutions. However, these methods do not consider robot dynamics or apply approximations that would sacrifice the modeling fidelity. For motion planning using full robot dynamics, such engineering compromise would limit the potential of the robots. % does not fully utilize the structure of the robotics configurations space. They

Thus, the natural question is how can we obtain the \emph{globally optimal} solution of motion planning problem at the dynamics level using \emph{exact} models? The main challenges are the scalability of the global optimization algorithm and the absence of proper robot formulations. For the former problem, recent progress in the Polynomial Optimization Problem (POP), i.e., Lasserre's hierarchy \cite{lasserre2001global, lasserre2015introduction}, enables one to compute globally optimal solutions of POPs via a sequence of Semidefinite Programming (SDP) that can be solved in polynomial time. For the latter problem, we introduce the Lie group-based formulation for robots composed of rigid bodies. 
\begin{figure}
\centering
\includegraphics[width=\columnwidth]{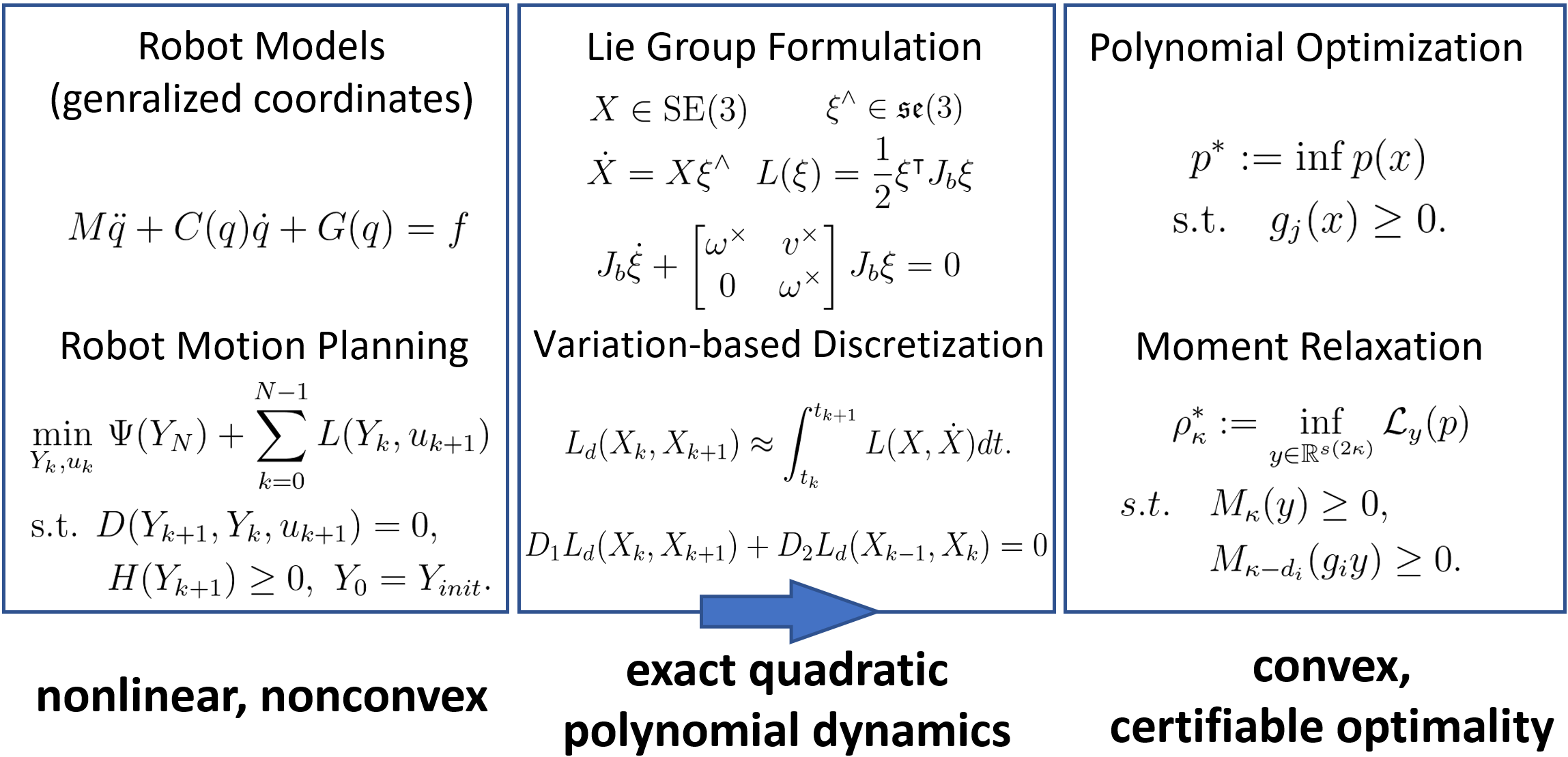}
\caption{The proposed geometric motion planning framework on Lie group. The motion planning problem is usually nonconvex due to the nonlinear dynamics. By utilizing robot models on Lie group and variation-based discretization, the motion planning problem can be converted to \emph{exact} quadratic polynomial optimization problem (POP). Via the moment relaxation method, i.e., Lasserre's hierarchy, the POP can be solved by a sequence of SDPs. We show that in our case, the second-order relaxation successfully provides certified globally optimal solutions for most of the testing cases despite the numerical challenges. }
\vspace{-20pt}
\end{figure}

In this paper, we show that bridging the geometric robotics formulation and Lasserre's hierarchy leads to certifiably optimal solutions for motion planning problems using the full dynamics model. We exploit the property of the configuration space of rigid body dynamics and apply variational integrator \cite{marsden2001discrete} on Lie groups to generate an exact polynomial formulation. Then we formulate the kinodynamic motion planning problem as a sparse POP with only quadratic polynomials. We further leverage Lasserre's hierarchy to approximate the globally optimal solutions. The main contributions of this paper are summarized as follows.
\begin{enumerate}
    \item Derivation of robotics dynamics model on matrix Lie groups using the variational integrator in discrete time. We show that the forced rigid body system can be represented by \emph{exact} quadratic polynomials.
    \item Formulation of kinodynamic motion planning problem as a low-order and sparse POP that can be solved via SDP at finite relaxation order. We show that the proposed formulation has linear time complexity with respect to the planning horizon. % The size of the resulting SDP is scalable for state-of-the-art SDP solvers.
    \item Evaluations of the proposed algorithms on 3D inverse kinematics for serial manipulators and trajectory planning for 3D drone using full dynamics model. We show that the second-order moment relaxations provide rank-one globally optimal solutions or provide infeasibility certificates for most of the testing cases modulo some numerical issues of the SDP solver. % \tsl{I replaced the tightness claim here.} % empirically tight to provide rank-one globally optimal solutions or provide infeasibility certificates. 
\end{enumerate}

\section{Related work}
% \tsl{Compare with search-based algorithm. Sampling based, MPPI for TO, not scalable. }
% \tsl{John Kelly's paper. Ashkan's paper}
%\mgj{Make it a separate section after the Introduction. It gets in the way of understanding the paper's merit, message, and contributions.}
\subsection{Sampling-based motion planning}
% Sampling-based motion planning methods are 
% p186 Lavalle, completeness. 
% The sampling-based motion planning methods has been demonstrated success in recent decades. The main idea of these methods is to avoid the explicit modeling of obstacles in the configurations space while using a sampling scheme to probe the nodes with collisions. These methods are complete or asymptotically optimal in the sense of probabilistic. Therefore, a solution is not guaranteed in finite time. 
The sampling-based motion planning algorithm has gained success in recent decades \cite{horsch1994motion,PRM,RRT,RRT-star,agha2014firm,hollinger2014sampling,ghaffari2019sampling}. These sampling-based methods are complete (or, resp. optimal), in the sense that the probability of finding the solution (or, resp. finding the optimal solution) converges to one when the sampling is enough \cite{lavalle2006planning, RRT-star}. However, as the completeness of these methods is only in the sense of probability, there is no guarantee of the runtime, and the algorithm may run forever if the solution does not exist. Due to the sampling nature of these algorithms, the solutions are chattering that need refining by local solvers.  

\subsection{Global optimization-based motion planning}
% GMP
% sparsity
% geometric perception
% some global optimization techniques effectively find or approximate the globally optimal solutions for motion planning problems. 

The mixed integer programming has been applied to the collision avoidance or path planning problem for aerial \cite{richards2002aircraft}, or ground vehicles \cite{schouwenaars2001mixed, deits2015efficient}, and legged robot \cite{ding2020kinodynamic, ding2018single, deits2014footstep}. The work of~\cite{richards2002aircraft, schouwenaars2001mixed, deits2015efficient} aims at path planning that omits the dynamics of the robots. The work of~\cite{ding2020kinodynamic, ding2018single, deits2014footstep} applied simplified dynamics models for legged robot step planning. These methods are based on simplified models, specifically for legged robots, and do not consider the 3D kinematics or dynamics constraints. Research \cite{dai2019global} represents the $\mathrm{SO}(3)$ surface by the convex hull in partitioned intervals, thus making the Inverse Kinematics (IK) problem a mixed-integer convex optimization. Then the branch-and-bound process \cite{lawler1966branch} is applied to solve the mixed-integer convex optimization. High accuracy approximation of the $\mathrm{SO}(3)$ would require more intervals that dramatically increase the runtime. These combinatorial methods are complete and capable of obtaining the globally optimal solutions \cite{lavalle2006planning}, while the cost is the exponential time complexity and omission of the geometry of the configuration space. The combinatorial methods cannot scale well for long-horizon planning and complex rigid body configurations. % The key of this work is to partition the convex hull of $\mathrm{SO}(3)$ into piece-wise linear intervals. 

Piece-wise linear kinematics model and Lasserre's hierarchy are applied in \cite{el2021piecewise} to plan paths with collisions defined by time-varying polynomial inequalities. With the increase of moment relaxation order, research \cite{el2021piecewise} could asymptotically find the collision-free path when a moving obstacle is presented. {Research \color{black}\cite{Jasour-RSS-21} extends such Lasserre’s hierarchy-based method to develop risk-bounded trajectory planners in the presence of uncertain time-varying obstacles using the notion of risk contours \cite{jasour2019risk}.} Moment relaxation methods have been applied to optimal control of hybrid systems \cite{zhao2019optimal} in continuous time. Via a sequence of SDPs, the control input and state monotonically converge to the global optimum. Sum-of-squares programming has been applied in \cite{tedrake2010lqr} to verify the region of attraction for the feedback controller.  Both \cite{zhao2019optimal} and \cite{tedrake2010lqr} consider Taylor expansions to approximate the nonlinear robot dynamics. As the Taylor expansion is an infinite series, finite order approximation can not be exact, and higher order approximation inevitably increases the size of the higher-order moment matrix. 
  
% To deal with realistic configuration space involved in IK problems, trigonometric functions have been represented by rational function in \cite{amice2023finding}, and \cite{trutman2022globally} to formulate 
Trigonometric functions have been represented as rational functions in \cite{amice2023finding,trutman2022globally} to formulate the kinematic constraints in the IK problem. Using this parameterization, \cite{trutman2022globally} applies the Lasserre's hierarchy to compute globally optimal solutions of IK and \cite{amice2023finding} apply the sum-of-squares \cite{parrilo2003semidefinite, blekherman2012semidefinite} to certify the collision-free regions. However, both \cite{amice2023finding} and \cite{trutman2022globally} only consider the rotation about a single axis and need to represent the joint $\mathrm{SE}(3)$ pose by a chain of trigonometric polynomials. The drawback is that when the kinematic chain increases, the degree of the polynomial representing the pose also increases. Additionally, \cite{amice2023finding, trutman2022globally} are dense formulations as each equality constraint involves all joint angles. The work of~\cite{maric2020inverse} casts the IK as a problem of finding the nearest point of an algebraic variety and proves all non-singular poses can be tightly solved by sparse SDP relaxation \cite{cifuentes2020local}. However, this formulation relies on distance geometry that only applies to kinematic problems. The work of~\cite{wang2020moment,jasour2021moment} describe the exact uncertainty propagation of systems parameterized by trigonometric polynomials using the notion of trigonometric moments. Though dynamics models are considered in both cases, the constraints between the Euler angles and their trigonometric functions are not linear. Such constraints make it hard to lift the dynamics constraints to the moment space. To solve the above issues, our geometric-based formulation is sparse and only involves quadratic polynomials that can be applied for long-horizon applications. % For motion planning with long horizons and dynamics models described by realistic configuration space, such dense and high-order formulations can not scale well. 

% For motion planning with long horizons and dynamics model with realistic configuration space, such dense and high-order formulations can not scale well. Our proposed geometric-based formulation is sparse and only involves quadratic polynomials that can be applied for long-horizon applications. 
% Trigonometric polynomials can accurately model the planar dynamics but have difficulty in describing the 3D full rotational motion. Euler angles are applied in \cite{jasour2021moment} to describe  3D rotation for aerial robot, but this has singularities and can result in higher-order polynomials up to six if discrete pose constraints are considered. 

% \tsl{I cited one more paper about the IK.} % A sparse sum-of-squares optimization has been applied to a distance geometry problem to solve the IK problems. 

% To deal with dynamic model, research \cite{wang2020moment}, and \cite{jasour2021moment} describe uncertainty propagation of systems in the form of trigonometric polynomials. Trigonometric polynomials can accurately model the planar dynamics but have difficulty in describing the 3D full rotational motion. Euler angles are applied in \cite{jasour2021moment} to describe  3D rotation for aerial robots, but this has singularities and can result in higher-order polynomials up to six if discrete pose constraints are considered. 

 \subsection{Lasserre's Hierarchy and certifiable optimality}
Lasserre's hierarchy converts the POP to infinite dimensional linear programming in measure space and approximates the global solution via truncated moment sequences in finite dimension \cite{lasserre2001global, lasserre2015introduction}. By increasing the order of the moment matrix, one can monotonically approximate the globally optimal solution of POP by solving a sequence of SDPs. When the cost function and constraints satisfy certain technical conditions \cite{nie2014optimality}, Lasserre's hierarchy can converge \emph{exactly} to the global optimum in finite relaxation order. For large-scale problems, the sparsity pattern in POP has been used to reduce the size of the SDP by breaking the dense moment matrix into smaller ones \cite{wang2021tssos, lasserre2006convergent, waki2006sums}. For control problems that satisfy Markov assumption, the sparsity pattern \cite{lasserre2006convergent, wang2022cs} makes the computation time linear in the problem horizon \cite{waki2006sums}. %  As the Lie group is a, there is a possibility that we can solve a class of control problems in a fixed relaxation order. could greatly accelerate the computation and

Lasserre's hierarchy has been applied to perception problem \cite{yang2022certifiably} as the certifiable algorithm \cite{bandeira2016note}. The certifiable algorithm \cite{bandeira2016note, yang2022certifiably} requires that: (i) the algorithm runs in polynomial time, (ii) returns a globally optimal solution with a certificate of the optimality, (iii) or fails to do so but provides a bound on the objective value. When the relaxation order of Lasserre's hierarchy is determined, the SDP can be solved in polynomial time. The optimum of SDP can also be certified as the globally optimal solution via rank conditions of the moment matrix or serve as a lower bound estimation for the problem \cite{lasserre2015introduction}. Such properties are rarely seen in the existing motion planning algorithms based on sampling, combinatorial solvers, or nonlinear local solvers. 

Continuous time control problems of polynomial systems have been lifted to space of measure to obtain optimal feedback \cite{henrion2008nonlinear, kamoutsi2017infinite, zhao2019optimal, yang2023suboptimal}, region of attraction \cite{henrion2013convex} or backward reachable set \cite{majumdar2014convex}. Though asymptotically convergent approximation by Lasserre's hierarchy is guaranteed, finite convergence is not observed in these cases. In this work, we leverage Lasserre's hierarchy by converting the discrete-time motion planning problem to exact POP for a single trajectory. We find that our numerical simulations can certify the optimal solution at the second relaxation order. % \tsl{Cited some paper about continuous-time formulation.}

\section{Background and Preliminary results} % Polynomial Optimization Problem and Lasserre's Hierarchy}
% In this section, we briefly review the polynomial optimization problem and introduce Lasserre's hierarchy. We mainly discuss Lasserre's hierarchy from the point of view of the general moment problem. 
\subsection{Polynomial optimization and Lasserre's Hierarchy}
Let $\mathbb{R}[x]$ be the ring of polynomial with real coefficients and $x:=(x_1, x_2, \dots, x_n)$. Given an integer $r$, we define the set $ \mathbb{N}_r^n := \{\alpha \in \mathbb{N}^n | \sum_i \alpha_i \le r \}$. The monomial with degree up to $r$ can be defined as $x^{\alpha}:=x_1^{\alpha_1}x_2^{\alpha_2}\dots x_n^{\alpha_n}, \alpha \in \mathbb{N}_r^n$, and we have the canonical basis for polynomial  degree up to $r$:
\begin{equation}
\label{eq:basis}
    v_r(x):= (1, x_1, x_2,\dots,x_n, x_1^2,x_1x_2,\dots,x_n^2,x_1^r,x_2^r,\dots,x_n^r). 
\end{equation}
Let $s(r):=\begin{pmatrix}
    n+r \\ 
    n
\end{pmatrix}$ be the dimension of $v_r(x)$. Then any $r$-degree polynomial $p(x):\mathbb{R}^n\rightarrow \mathbb{R}$ could be expressed as:
\begin{equation}
    p(x)=\sum_{\alpha}p_{\alpha}x^{\alpha} = \langle p, v_r(x)\rangle ,
\end{equation}
where $p=\{p_{\alpha}\}$ denotes the coefficients corresponding to the basis defined in \eqref{eq:basis}.  
We define the POP as follows.
\begin{problem}[Polynomial Optimization Problem]
\begin{equation}
\label{prob:pop}
\begin{aligned}
    p^*&:=\inf p(x) \\
    \text{s.t.} \quad &g_j(x) \ge 0, \quad \forall j \in \{1,\ldots,m\}. \\
    %\quad &h_i(x) = 0, i = 1,\dots,l_h \\
\end{aligned}\tag{POP}
\end{equation}
where $p$, $g_i$ are polynomials. We denote the feasible set as $\mathbb{K}$. % We assume that $\mathbb{K}$ is compact. % \tsl{Add equality constraints or not.}
\end{problem}
We denote the degree, i.e., the highest order of monomial, of polynomial $g$ as $\operatorname{deg}g$. Then we have the degree integers as $d_i := \left\lceil \operatorname{deg}(g_i) / 2 \right\rceil$ and $d_g = \max\{1, d_1,\dots,d_m\}$, where $\left\lceil a \right\rceil$ denotes the integer greater than or equal to $a$. % TODO: How to use half a page to make it clear? 

Given a probability distribution $\mu(x)$ in $\mathbb{R}^n$ and $\alpha \in \mathbb{N}_r^n$, then the moment of $\mu(x)$ at order $\alpha$ is defined as:
\begin{equation}
    y_{\alpha}=y_{\alpha_1,\dots, \alpha_n}=\mathbb{E}[x^{\alpha}]=\int x^{\alpha}\mu(x)dx .
\end{equation}
We construct the (truncated) moment matrix $M_r(y) \in \mathbb{R}^{s(r)\times s(r)}$ via a $s(2r)$-sequence $y=(y_{\alpha})$, with rows and columns labeled the same as in \eqref{eq:basis}. For example, moment matrix  $M_2(y)$, with $n=2, r=2$, is:
\begin{equation}
    {M}_{2}(y) =\left[\begin{array}{cccccc}
y_{0,0} & y_{1,0} & y_{0, 1} & y_{2,0} & y_{1,1} & y_{0,2} \\
y_{1,0} & y_{2,0} & y_{1, 1} & y_{3,0} & y_{2,1} & y_{1,2} \\
y_{0,1} & y_{1,1} & y_{0, 2} & y_{2,1} & y_{1,2} & y_{0,3} \\
y_{2,0} & y_{3,0} & y_{2, 1} & y_{4,0} & y_{3,1} & y_{2,2} \\
y_{1,1} & y_{2,1} & y_{1, 2} & y_{3,1} & y_{2,2} & y_{1,3} \\
y_{0,2} & y_{1,2} & y_{0, 3} & y_{2,2} & y_{1,3} & y_{0,4}
\end{array}\right]
\end{equation}
Suppose $y = (y_{\alpha}) \subset \mathbb{R}$ be a sequence indexed by 
$\alpha \in \mathbb{N}_r^n$, and let the $\mathcal{L}_y: \mathbb{R}[x] \rightarrow \mathbb{R}$ be the linear functional:
\begin{equation}
\label{eq:functional}
    f(x) = \sum_{\alpha} f_{\alpha} x^{\alpha} \mapsto \mathcal{L}_y(f) = \sum_{\alpha} f_{\alpha}y_{\alpha}. % \mathbb{N}^{m}
\end{equation}
The functional \eqref{eq:functional} can be interpreted as substituting the monomials $x^{\alpha}$ in $f(x)$ by corresponding $y_\alpha$ to obtain the numerical value. Then $M_r(y)$ can be constructed by: % \textbf{}
\begin{equation}
\label{eq:moment_matrix}
    M_r(y)(\alpha, \beta) = \mathcal{L}_y(x^{\alpha} x^{\beta}) = y_{\alpha + \beta}, \alpha, \beta \in \mathbb{N}_r^n,
\end{equation}
or equivalently, by manipulating the entire $v_r(x)v_r(x)^{\transpose}$:
\begin{equation}
    M_r(y) = \mathcal{L}_y(v_r(x)v_r(x)^{\transpose}). 
\end{equation}
Finally, we also have the localizing matrix for $g_i$ as:
\begin{equation}
    M_{r - d_i}(g_iy) = \mathcal{L}_y(g_iv_{r-d_i}(x)v_{r-d_i}(x)^{\transpose}). 
\end{equation}
% \subsection{Lasserre's Hierarchy}
One equivalent formulation of \eqref{prob:pop} is infinite-dimensional linear programming over the space of probability with support on the feasible set $\mathbb{K}$ \cite{lasserre2001global, lasserre2015introduction}, where the objective function is a linear combination of the entries of the moment matrix. The detailed formulation of the infinite-dimensional linear programming formulation is presented in Appendix~\ref{appx:inf-lp}.   

As searching over an infinitely large moment matrix is impossible, we approximate the solution by sequences $y_{\alpha}$ with finite order $|\alpha| = 2\kappa$. Thus, we have the relaxed SDP in the space of moment matrix $M_{\kappa}(y)$ as follows.
\begin{problem}[Semidefinite relaxation of~\ref{prob:pop}]
\begin{equation}
\begin{aligned}
\label{prob:pop_sdp}
    % \rho_N^*&:=\inf \sum_{\alpha} p_{\alpha}y_{\alpha} \\
    \rho_{\kappa}^*&:=\inf_{y\in \mathbb{R}^{s(2\kappa)}} \mathcal{L}_y(p) \\
    s.t. \quad &M_{\kappa}(y) \ge 0, \\
    \quad &M_{\kappa-d_i}(g_iy) \ge 0, \forall j \in \{1,\ldots,m\}.
\end{aligned}\tag{SDP}
\end{equation}
where $p$ and $g_j$ are polynomials. 
\end{problem}
% \eqref{prob:pop_sdp} is a convex optimization problem that can be solved in polynomial time when the order $\kappa$ is determined. Assuming that the minimizer of \eqref{prob:pop_sdp} is $y^*$, the remaining issue is to verify that the $y^*$ indeed admits a representing measure over $\mathbb{K}$. For a sequence with finite order, we have the following sufficient condition. 

Because Problem~\eqref{prob:pop_sdp} gets more constrained as $\kappa$ increases, we could gradually approximate the globally optimal solution of \eqref{prob:pop}. This observation leads to the theory of Lasserre's hierarchy.
\begin{theorem}[Lasserre's Hierarchy \cite{lasserre2001global, lasserre2015introduction}]Let $p^*$ be the optimum of \eqref{prob:pop} and the $\rho_{\kappa}^*$ (resp. $y^*_{\kappa}$) be the optimum (resp. optimizer) of \eqref{prob:pop_sdp}, then: 
\begin{enumerate}
    \item (Monotone lower bound) $\rho_{\kappa}^*$ is monotonically increasing and $\rho_{\kappa}^*\uparrow p^*$ as ${\kappa} \rightarrow \infty$.
    \item (Rank condition) If the moment matrix satisfies: $$\operatorname{rank}(M_{\kappa}(y_{\kappa}^{*})) = \operatorname{rank}(M_{{\kappa} - d_g}(y_{\kappa}^{*})),$$ then $\rho_{\kappa}^* = p^*$. In this case, $y^*$ is a moment sequence that admits a representing measure on $\mathbb{K}$. % Otherwise, one can increase $\kappa$ and solve \eqref{prob:pop_sdp} until the rank condition is satisfied. % Note that the rank condition is necessary for $y_{\kappa}^*$ to have a representing measure over $\mathbb{K}$.
    \item (Number of optimizers) If 2) is satisfied, then the number of optimizers equals to $\operatorname{rank}(M_{\kappa}(y_{\kappa}^{*}))$. 
    \item (Finite convergence) If \eqref{prob:pop} satisfy some suitable technical condition, under the assumption that Archimedeanness condition holds for $\mathbb{K}$, then $\rho_{\kappa}^* = p^*$ happens at some finite order ${\kappa}^*$ \cite{nie2014optimality}. Note that the actual order ${\kappa}^*$ is unknown in advance. 
\end{enumerate}
\end{theorem}
Lasserre's hierarchy indicates that one can try from the SDP relaxation with the lowest order and test the rank condition until it is satisfied to obtain the globally optimal solution. For a wide range of applications that only have one optimal solution \cite{yang2020one, yang2022certifiably}, the rank one optimality condition is expected using Lasserre's hierarchy. 
\begin{remark}[Rank one optimality condition]
    If $\operatorname{rank}(M_{\kappa}(y_{\kappa}^{*})) = 1$, then $\operatorname{rank}(M_{{\kappa} - d_g}(y_{\kappa}^{*})) = 1$ as $M_{{\kappa} - d_g}(y_{\kappa}^{*})$ is a non-zero principle submatrix of $M_{\kappa}(y_{\kappa}^{*})$. 
\end{remark}
% For sparse moment relaxation problems, we refer the reader to Appendix. XX. 
\subsection{Rigid body dynamics}
We consider special Euclidean group $\mathrm{SE}(3)$ as the configuration space of rigid body motion, where the state \mbox{$X = \begin{bmatrix}
    R & p \\ 0 & 1
    \end{bmatrix} \in \mathrm{SE}(3)$} can be represented by the orientation
% rotation matrix $R$ in the special orthogonal group $\mathrm{SO}(3)$:
$$R \in \mathrm{SO}(3) = \{R\in \mathbb{R}^{3 \times 3} \mid R^{\transpose}R = I_3, \det(R) = 1\},$$ and position $p \in \mathbb{R}^3$. 
% The homogeneous representation of an element in $\mathrm{SE}(3)$ is given by
% \begin{equation}
%     X = \begin{bmatrix}
%     R & p \\ 0 & 1
%     \end{bmatrix} \in \mathrm{SE}(3).
% \end{equation}
On $\mathrm{SE}(3)$, the twist is defined as the concatenation of linear velocity $v$ and angular velocity $\omega$ in the body frame, i.e., $$\xi := \begin{bmatrix} \omega \\ v \end{bmatrix} \in \mathbb{R}^6, 
\xi^{\wedge}=\begin{bmatrix} \omega^\times & v \\ 0 & 0 \end{bmatrix}\in\mathfrak{se}(3),$$ where $(\cdot)^{\times}$ satisfies $a^{\times}b = a\times b, a,b\in \mathbb{R}^3$. Note that $\mathfrak{se}(3)$ is the tangent space at the identity $X=I$, and $X\xi^{\wedge} \in T_X \mathrm{SE}(3)$. 
The reconstruction equation gives the Equation of Motion (EOM) in continuous time:
\begin{equation}
\label{eq:reconstruction}
    \dot{X}=X\xi^{\wedge}.  % \text{ i.e., } \dot{R} = R\omega^{\wedge},\ \ \ \dot{p} = Rv
\end{equation}
We have the inertia matrix $J_b$ and the kinetic energy:
\begin{equation}
    \label{eq:inertia_matrix}
    J_b := \begin{bmatrix} I_b & 0 \\ 0 & mI_3 \end{bmatrix},\ \ T(\xi):=\frac{1}{2}\xi^{\transpose}J_b\xi ,
\end{equation}
where $I_b \in \mathbb{R}^{3\times 3}$ is the moment of inertia in the body frame and $m$ is the body mass. We arrive at the Euler-Poincar\'e equation \cite{bloch2003nonholonomic, marsden1998introduction} if we take a variation in $T_X \mathrm{SE}(3)$:
\begin{equation}
    \label{eq:rigid_body_dynamics}
     J_b \dot{\xi}  + \begin{bmatrix}
    \omega^\times & v^\times \\ 0 & \omega^\times
    \end{bmatrix} J_b \xi = 0. 
\end{equation}

The challenge of applying Lasserre's hierarchy is to represent the rigid body motion via polynomials. Although one can apply conventional integration schemes such as explicit Euler or Runge-Kutta method on \eqref{eq:rigid_body_dynamics} in vector space, the integration of the continuous time EOM \eqref{eq:reconstruction} on $\mathrm{SE}(3)$ involves the exponential map that does not have polynomial formulations in finite order \cite{celledoni2014introduction, iserles2000lie}. If one uses Euler angles to parameterize $R \in \mathrm{SO}(3)$, then a mixture of trigonometric polynomial and $\mathbb{R}[x]$ is expected, which highly increases the complexity \cite{jasour2021moment}. Instead, we implement the Lie Group Variational Integrator (LGVI) that approximates the generalized velocity via the configuration state to avoid the issue of integrating the continuous time EOM~\cite{marsden2001discrete}. 

\subsection{Variational integrator}
\label{sec:VI_prelim}
% \tsl{Need a remark about the nonholonomic constraints.} $n$-dimensional 
Consider a mechanical system with the configuration space $Q$. We denote the configuration state as $q \in Q$ and the generalized velocity as $\dot{q} \in T_{q}Q$. Then we have the Lagrangian given the kinetic and potential energy $T(\dot{q}), V(q)$:
\begin{equation}
\label{eq:ct_lag}
    L(q, \dot{q}):=T(\dot{q}) - V(q). 
\end{equation}
The key idea of a variational integrator is to discretize the Lagrangian \eqref{eq:ct_lag} to obtain the discrete-time EOM \cite{marsden2001discrete}. The discretization scheme ensures that the Lagrangian is conserved in the discrete-time, thus having superior energy conservation property in the long duration. 

We define the time step $h \in \mathbb{R}$ and the time sequence \mbox{$\{t_k = kh \mid  k = 0, \dots, N\} \subset \mathbb{R}$}. Thus the discrete Lagrangian $L_d: Q\times Q \rightarrow \mathbb{R}$ could be considered as the approximation of the action integral via
\begin{equation}
\label{eq:lag_ct2dt}
    L_d(q_k, q_{k+1}) \approx \int ^{t_{k+1}}_{t_k}L(q, \dot{q})dt .
\end{equation}
In this work, we consider the midpoint approximation \cite{marsden2001discrete}:
\begin{equation}
\label{eq:dt_lag_int}
    L_d(q_k, q_{k+1}) = hT(\frac{q_{k+1} - q_k}{h}) - hV(\frac{q_{k+1} + q_k}{2}).
\end{equation}
Then the discrete variant of the action integration becomes:
\begin{equation}
    S_d = \sum_{k=0}^{N-1}{L}_d(q_k, q_{k+1}) .
\end{equation}
% where $\bar{L}_d(q_k, q_{k+1})$ is the sum of the midpoint approximation defined in \eqref{eq:dt_lag_int} and \eqref{eq:dt_force_int}. 
Finally, we take variation in $TQ$ and group the term corresponding to $\delta q_{k} \in T_{q_k}Q$ as the discrete version of integration by parts \cite{marsden2001discrete}: 
\begin{equation}
\begin{aligned}
  \delta S_d &= D_1 {L}_d(q_0, q_1) \cdot \delta q_0 +  D_2 {L}_d(q_{N-1}, q_N) \cdot \delta q_N\\ 
    &+ \sum_{k=1}^{N-1}
    \left( D_2 {L}_d(q_{k-1}, q_k) + D_1 {L}_d(q_k, q_{k+1})\right) \cdot \delta q_k. 
\end{aligned}
\end{equation}
where $D_i$ denotes the derivative with respect to the $i$-{th} argument. 
By the least action principle, the stationary point can be derived: 
\begin{equation}
\label{eq:discrete_eom}
D_1L_d(q_k, q_{k+1}) + D_2L_d(q_{k-1}, q_{k}) = 0. % Bf_d(t_k)
\end{equation}
To incorporate the external force $f \in T^*_qQ$, we can compute the action integral again using the midpoint approximation as:
\begin{equation}
\label{eq:dt_force_int}
    \int_{t_k}^{t_{k+1}}f(t)\cdot\delta q \approx \frac{h}{2}f(t_k)\cdot \delta q_{k} + \frac{h}{2} f(t_{k+1})\cdot \delta q_{k+1},
\end{equation}
and then incorporate it into~\eqref{eq:discrete_eom}. We will later show that discretizing Lagrangian on $\mathrm{SE}(3)$ results in polynomial dynamics.

% By variational integrator, all the velocity $\dot{q}$ are represented by configuration state $q$.% Considering the configuration states of $\mathrm{SE}(3)$ are polynomials, discretizing the Lagrangian creates a polynomial which enables us to apply Lasserre's hierarchy.

\section{Method and Main Results}
This section introduces our method to formulate the kinodynamic motion planning problem as exact POP using matrix Lie group formulations. 

\subsection{Problem formulation}
In what follows, we consider the discrete-time kinodynamic motion planning problem of a rigid body system.
\begin{problem}[Discrete Kinodynamic Motion Planning]
\label{prob:kindyn_mp}
    \begin{equation}
    \begin{aligned}
    \label{prob:pocp}
    \min_{Y_k, u_k} & \ \Psi(Y_N) + \sum_{k=0}^{N-1} L(Y_k, u_{k+1}) \\ 
    \text{s.t. } & D(Y_{k+1}, Y_k, u_{k+1}) = 0, \ k = 0, \dots, N-1, \\
    & u_{\min} \le u_k \le u_{\max}, \ H(Y_{k+1}) \ge 0, \ Y_0 = Y_{\text{init}}.
    % & \quad Y_0 = Y_{\text{init}}. \\
    \end{aligned}
\end{equation}
\end{problem}
We require that the system state $Y$ contains the configuration state $X \in \mathrm{SE}(3)$ and velocity $\xi \in \mathfrak{se}(3)$ or its discrete variants. $D(\cdot, \cdot)$ is the EOM subject to the initial condition $Y_0 = Y_{\text{init}}$. For a rigid body system, we consider the {\color{black} discretized} motion by \eqref{eq:reconstruction} and \eqref{eq:rigid_body_dynamics}. $u_{\max}$ and $u_{\min}$ are the input constraints, and $H(Y)\ge0$ denotes other constraints, including constraints of $\mathrm{SO}(3)$ manifold and obstacle avoidance. $\Psi(\cdot)$ is the terminal cost, and $L(\cdot, \cdot)$ is the stage cost. If we consider the terminal constraint, we replace $\Psi(\cdot)$ with the equality constraint $Y_N=Y_{g}$ considering the terminal state $Y_g$.

To apply Lasserre's hierarchy, we require that all the costs and constraints in Problem~\ref{prob:kindyn_mp} are polynomials. As we applied a direct collocation \cite{betts1998survey,hereid2017frost} style formulation, the order of polynomial will not grow as the planning horizon increases. % and the order should be as low as possible
\subsection{Polynomial dynamics constraints }
We now derive the dynamics model on $\mathrm{SE}(3)$ via LGVI~\cite{marsden1999discrete, lee2005lie,lee2007lie,nordkvist2010lie}. The derivation for on $\mathrm{SO}(3)$ has been well-established in \cite{marsden1999discrete, lee2005lie,lee2007lie}. However, a formulation on $\mathrm{SE}(3)$ suitable for POP implementation is absent. We apply the midpoint approximation \eqref{eq:dt_lag_int} to represent the twist $\xi_k$ using the configuration state:
%Then, we the midpoint of $\xi_k$ using $X_k$ and $X_{k+1}$: 
\begin{equation}
\label{eq:Fk_midpoint}
\begin{aligned}
    F_k := R_k^{-1}R_{k+1} \approx I + h\omega_k^{\times}, \ \ \omega_k^{\times} \approx \frac{F_k - I}{h},
\end{aligned}
\end{equation}
\begin{equation}
\label{eq:pk_midpoint}
\begin{aligned}
    \dot{p}_k = R_kv_k \approx \frac{p_{k+1} - p_k}{h},\ \ v_k \approx \frac{R_k^{\transpose}(p_{k+1} - p_k)}{h},
\end{aligned}
\end{equation}
with $F_k \in \mathrm{SO}(3)$ the pose change. 
We refer to \cite{lee2005lie} for the expression of the kinetic energy of rotation. Then by \eqref{eq:dt_lag_int}, the discrete kinetic and potential energy takes the form:
\begin{align}
    \nonumber T_d&:=\frac{1}{2h}\operatorname{tr}((F_k - I)I^b(F_k - I)^{\transpose}) + \frac{1}{2h}m\|p_{k+1} - p_k\|^2, \\
    V_d&:=h m\frac{p_{k+1}+p_{k}}{2}\cdot g ,
\end{align}
where $g \in \mathbb{R}^3$ is the gravity and $I^b$ the nonstandard moment of inertia \cite{marsden1999discrete} that relate the standard moment of inertia $I_b$ by $I_b = tr(I^b)I_3-I^b$. Then we define the variation $\delta X \in T_X \mathrm{SE}(3)$ as:
\begin{equation}
    \delta X = X \delta\eta^{\wedge} \in T_X \mathrm{SE}(3), \delta\eta^{\wedge}=\begin{bmatrix} \delta \omega^{\times} & \delta \rho\\ 0& 1 \end{bmatrix} \in \mathfrak{se}(3). 
\end{equation}
The detailed derivation and the explanation of the $I^b$ are well presented in \cite{marsden1999discrete, lee2005lie}. Here we only derive the position part not presented in the existing literature. Consider $V_d$ and 
$$T_{d,p}(p_k, p_{k+1}) = \frac{1}{2h}m\|p_{k+1} - p_k\|^2,$$
then we have the variation of position in the world frame as $\delta p = R\delta \rho$. 
Using \eqref{eq:discrete_eom}, we have:
\begin{equation}
\label{eq:discrete_T}
    \begin{aligned}
        D_1T_{d, p}  &= m(\frac{p_k - p_{k+1}}{h})\cdot R_k \delta \rho_k = m(R_k^{\transpose}\frac{p_k - p_{k+1}}{h}) \cdot  \delta \rho_k \\
        D_2T_{d, p}  &= m(\frac{p_{k} - p_{k-1}}{h})\cdot R_k \delta \rho_k = m(R_k^{\transpose} \frac{p_{k} - p_{k-1}}{h}) \cdot \delta \rho_k \\
    \end{aligned}
\end{equation}
% \begin{equation}
% \label{eq:discrete_T}
%     \begin{aligned}
%         D_1T_{d, p}  &= m(- \frac{p_{k+1} - p_k}{h})\cdot R_k \delta \rho_k \\
%         & = m(- R_k^{\transpose}\frac{p_{k+1} - p_k}{h}) \cdot  \delta \rho_k \\
%         D_2T_{d, p}  &= m(\frac{p_{k} - p_{k-1}}{h})\cdot R_k \delta \rho_k \\
%         & = m(R_k^{\transpose} \frac{p_{k} - p_{k-1}}{h}) \cdot \delta \rho_k \\
%     \end{aligned}
% \end{equation}
Similarly, we have the variation for the potential energy as
\begin{equation}
\label{eq:discrete_V}
    \begin{aligned}
        D_1V_{d} = D_2V_{d} = \frac{hmg}{2}\cdot  R_k \delta \rho_k = R_k^{\transpose}\frac{hmg}{2}\cdot \delta \rho_k
    \end{aligned}
\end{equation}
We wrap up \eqref{eq:discrete_T} and \eqref{eq:discrete_V} to obtain 
\begin{equation}
\label{eq:lgvi_pos}
    mR_k^{\transpose}\frac{p_{k+1} - p_k}{h} = mR_k^{\transpose}\frac{p_{k} - p_{k-1}}{h} + R_k^{\transpose}mg h. 
\end{equation}
Then, by substituting \eqref{eq:Fk_midpoint} and \eqref{eq:pk_midpoint} into \eqref{eq:lgvi_pos}, we get
\begin{equation}
    mv_{k+1} = mF^{\intercal}_kv_{k} + h mR_{k+1}g. 
\end{equation}
We now present the LGVI for the unconstrained rigid body:
\begin{equation}
\label{eq:lgvi_dynamics}
\begin{aligned}
    &I^bF_{k+1}^{\transpose} - F_{k+1}I^b=F_k^{\transpose}I^b - I^bF_{k}, \\
    &mv_{k+1} = mF^{\intercal}_kv_{k} + h mR_{k+1}g .
    % & s.t. F_k, R_k, R_{k+1}, F_{k+1}\in \mathrm{SO}(3)
\end{aligned}
\end{equation}

\begin{figure}[t]
    \centering
    \includegraphics[width=1\columnwidth]{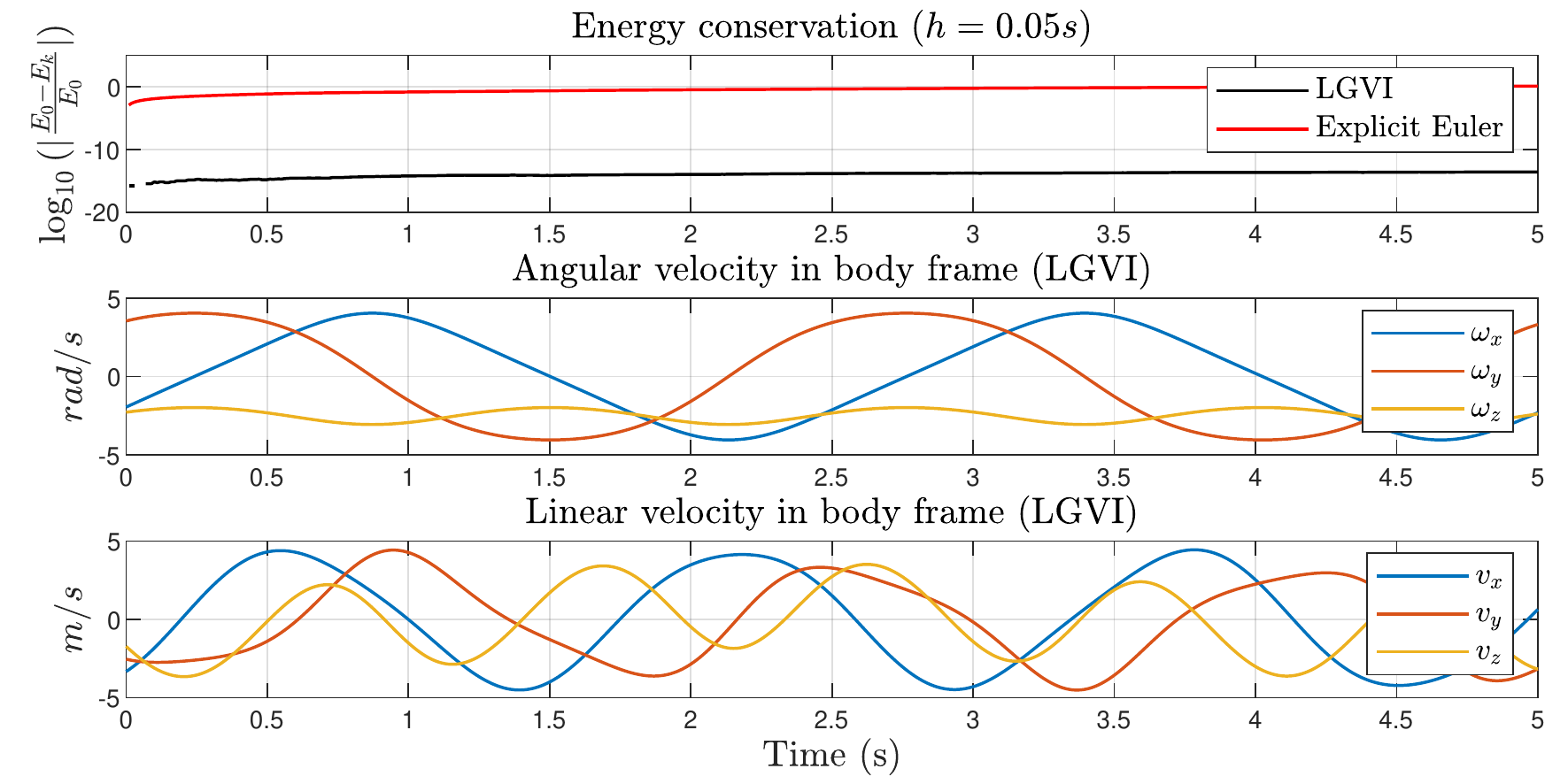}
    \caption{Comparison of the proposed LGVI with explicit 
    Euler integrator for the rigid body system. The presented twists are computed by LGVI. Considering the energy at time step $k$, the normalized energy loss $\frac{|E_k - E_0|}{E_0}$ is negligible for LGVI while the explicit Euler integrator soon diverges.} % \tsl{Correct the legends.}
    \label{fig:LGVI_compare}
    % \vspace{-20pt}
\end{figure}

% Consider the torque $\tau\in \mathbb{R}^3$ and force $f\in\mathbb{R}^3$ applied in the body frame, we have the action integral:
% \begin{equation}
%     \frac{h}{2} \int_{t_k}^{t_{k+1}} R_{k}\tau_k \cdot R_k\delta \rho_k
% \end{equation}
We note that all constraints are polynomials with degrees up to two. Similar derivation for $\mathrm{SE}(3)$ has been seen in the research \cite{nordkvist2010lie, lee2007lie}. However, the presented derivation directly applies variations on $T{\mathrm{SE}(3)}$ in discrete time and does not rely on the continuous angular velocity $\omega$ as in~\cite{nordkvist2010lie}. The elimination of $\omega$ is more suitable for controller design as it is redundant if we have its discrete variant $F_k$. The method in \cite{lee2007lie} does not consider $v_k$ in the tangent space, which is unsuitable for controller design. To verify the correctness of the derived integrator, we plot the kinetic energy and twist for the system without gravity in Fig.~\ref{fig:LGVI_compare}. We can see that the kinetic energy is conserved in the long run, while the popular explicit Euler method with dynamics \eqref{eq:rigid_body_dynamics} diverges fast. % Though one can represent the dynamics completely using configuration state \cite{lee2007lie}, we find that reserving $v_k$ and $F_k$ leads to a sparser moment relaxation. % The detailed derivation is introduced in the appendix.

\subsection{Polynomial kinematics constraints}
% Consider $X_k, X_{k+1} \in \mathrm{SE}(3)$, we define the pose change as the $\Delta X_k := \begin{bmatrix}F_k & \Delta p_k \\ 0 & 1 \end{bmatrix}$, with $F_k:= $
We consider the midpoint approximation as our discrete-time equation of motion for the kinematic part:
\begin{equation}
\label{eq:lgvi_kinematic}
    R_{k+1} = R_kF_k,\ \ p_{k+1} = hR_kv_k + p_k,
\end{equation}
which are also quadratic polynomials. As $R \in \mathrm{SO}(3)$ contains nine entries while $\dim \mathrm{SO}(3) = 3$, we need additional constraints. Consider the column space:
$$R:= [r_1, r_2, r_3]\in \mathbb{R}^{3\times 3}, \quad r_1,r_2,r_3\in \mathbb{R}^3.$$ $R\in \mathrm{SO}(3)$ is equivalent to the following 15 quadratic equality constraints: % to impose $R$ on $\mathrm{SO}(3)$ manifold
\begin{equation}
\label{eq:SO3_cons}
\begin{aligned}
    &\|r_1\|^2-1=\|r_2\|^2-1=\|r_3\|^2-1 = 0,\\
    & r_1^{\transpose}r_2 = r_2^{\transpose}r_3 = r_1^{\transpose}r_3 = 0,\\
    &r_1\times r_2 - r_3 = r_2\times r_3 - r_1 = r_3\times r_1 - r_2 = 0_{3\times 1}.
\end{aligned}
\end{equation}
The first six equations ensure that $r_i$ are orthonormal to each other, and the last nine equations ensure $r_i$ follows the right-hand rule, which is equivalent to the determinant constraints but is quadratic. 

We introduce two variables, $c$ and $s$ that satisfy $c^2 + s^2 = 1$ to indicate rotation about a single axis:
\begin{equation}
\small 
\label{eq:joint_rotate}
\begin{aligned}
    &R_x = \begin{bmatrix} 1 & 0 & 0 \\ 0 & c & -s \\ 0 & s & c\end{bmatrix}, 
    R_y = \begin{bmatrix} c & 0 & s \\ 0 & 1 & 0 \\ -s & 0 & c\end{bmatrix}, R_z = \begin{bmatrix} c & -s & 0 \\ s & c & 0 \\ 0 & 0 & 1\end{bmatrix}.
\end{aligned}
\end{equation}
$R_{i}(\theta)$ denotes a rotation of $\theta$ angle about axis $i$. 

\subsection{Sparsity pattern}
As Problem \ref{prob:kindyn_mp} satisfies the Markov assumption, we show it enables the use of Correlative Sparsity (CS) that greatly reduces the computational burden~\cite{lasserre2006convergent}. Consider the full dynamics model represented in \eqref{eq:lgvi_dynamics} and \eqref{eq:lgvi_kinematic}, we can formulate $Y$ in Problem \ref{prob:kindyn_mp} as $Y_k := \left(R_k, p_k, F_k, v_k\right)$. Then we also include  \eqref{eq:lgvi_dynamics}-\eqref{eq:SO3_cons} in constraints $D$ and $H$. We partition the state into $N$ sets $y(I_k) = \{(Y_{i}, Y_{i-1}, u_i)| i \in I_k\}$ indexed by $I_k = \{k, k+1\}, k=1,2,\dots, N$. Then we can verify the running intersection property \cite{lasserre2006convergent}: $$\forall k=1,\dots, q-1, \exists s \le k, I_{k+1} \cap \left(\cup_{j=1}^kI_j\right) \subseteq I_s,$$ by selecting $s=k$, i.e. $I_{k+1} \cap \left(\cup_{j=1}^kI_j\right) = \{k+1, k+2\} \cap \left(\cup_{j=1}^kI_j\right) = \{k+1\} \in I_k$. We can see that the cost and dynamic constraints can also be partitioned into such indexed sets as $D(Y_{k+1}, Y_k, u_{k+1}) = 0$, $H(Y_{k+1}) = 0$ and $L(Y_k, u_{k})$ only involves variable $y(I_{k})$. These properties enable the sparse moment relaxation to have the same result as the dense version if one breaks the dense matrix $M_{\kappa}(y)$ to $N$ smaller ones that only involve variables in $y(I_k)$ \cite{lasserre2006convergent}. The detail of CS is presented in Appendix~\ref{appx:sparse-pop}. When the relaxation order is determined, we can see that the size of the variable grows linearly with respect to the planning horizon \cite{lasserre2006convergent,lasserre2015introduction}. 

Other than CS, Term Sparsity (TS) has also been exploited to reduce the computational burden further~\cite{wang2021tssos, yang2022certifiably} by eliminating more variables in the moment matrix. Though relaxation with TS is looser than CS, the computation time is greatly reduced.

%, which is impossible if we apply the dense formulation as in \cite{amice2023finding, trutman2022globally}. %To illustrate the sparsity pattern, we will show ta

% Thus, the number of moment matrix grows linearly with respect to the planning time horizon N
\section{Numerical simulations}
In this section, we apply the proposed framework to various robotics applications. 

\subsection{Simulation setup}

% \subsubsection{Matrix calculations}
\subsubsection{Relaxation scheme}
% We explicitly model all the robot state as well as the control input to ensure the correlative sparsity pattern is available. 
For all tasks, the dynamics, constraints, and objective functions are quadratic polynomial functions; thus, we have the lowest order for Lasserre's hierarchy $\kappa = 1$ and the degree parameter $d_g=1$. However, $\kappa = 1$ only returns a trivial lower bound with infeasible solutions. Thus we try $\kappa = 2$ and find it works well for our cases. % verified principle submatrix $M_{\kappa - d_g}$ to

We use the recent state-of-the-art tool CS-TSSOS \cite{wang2021chordal, wang2022cs, wang2021tssos, magron2021tssos} to explore the CS and TS pattern. Chordal extension \cite{wang2021chordal} is applied to boost both CS and TS by either extending matrix size (for CS) or reducing more terms (for TS). As the CS does not sacrifice the tightness and the problem size is huge, we always exploit the CS pattern. The CS-TSSOS supports the maximal or approximately smallest chordal extension, which is denoted as MD and block, respectively, referring to the programming API. Assume that the global optimum of \eqref{prob:pop} is $p^*$ and the optimum of the \eqref{prob:pop_sdp} or local Nonlinear Programming (NLP) solver is $\rho$, then we have the following inequalities:
\begin{equation}
\begin{aligned}
    \rho_{TS+MD} \le \rho_{TS+block} &\le \rho_{CS} \le \rho_{CS+MD} \\
   \le \rho_{dense} &\le p^* \le \rho_{NLP},
\end{aligned}
  % \rho_{_{TS+MD}} \le \rho_{_{TS+block}} \le \rho_{_{CS}} \le  \rho_{_{CS+MD}} \le \rho_{_{dense}} \le p^* \le \rho_{_{NLP}},
\end{equation}
where the subscripts denote different sparsity patterns at the same order $\kappa$. As the NLP solver are based on local gradient information, only local optimum is guaranteed. Therefore, $\rho_{NLP}$ serves as an upper bound of $p^*$.  % $\rho_{NLP}$ is a local optimum obtained via any . 
\subsubsection{Evaluation metric}
As the optimal value of \eqref{prob:pop} can not be greater than $\rho_{NLP}$, then the problem is upper and lower bounded by the SDP and NLP result. Thus, we can use the relative suboptimality as an index of the relaxation gap:
\begin{equation}
    \epsilon := \frac{\rho_{NLP} - \rho_{SDP}}{ \rho_{NLP}}. \tag{suboptimality}
\end{equation}

As the rank condition is subject to numerical error, we instead check the eigenvalues of each sub-moment matrix as is shown in Remark~\ref{remark:sparse_rank_one} in Appendix~\ref{appx:sparse-pop}. We check if each moment matrix is rank-one, assuming the unique minimizer. Assume the eigenvalues of $k$-th moment matrix $\lambda_{k, i}$ are ranked by their $|\lambda_{k,i}|$ in descending order. Then we compute the ratio between the first and the second one to represent the rank condition: % $M_{\kappa}(y, I_k)$
\begin{equation}
    \delta_k = \frac{|\lambda_{k, 2}|}{|\lambda_{k, 1}|} \le 1, \delta = \max_{k} \delta_k. \tag{rank condition}
\end{equation}
As the numerical accuracy is still the bottleneck of the current SDP solvers for problems of this size, $\delta$ will only be used for comparison with $\epsilon$. % and will not be used as a threshold to decide if the rank-one condition is satisfied. 
% At relaxation order $\kappa = 2$, we find that the SDP either return a certified optimal value or a non-trivial lower bound estimation of the optimum in all the tested case. For the latter case, we find that the solution can be a good warm start for general nonlinear solvers that converge after a few function evaluations. 

\subsubsection{Software and hardware setup}
The CS-TSSOS is implemented in Julia's open-source package \cite{wang2022cs, wang2021tssos}. CS-TSSOS converts \eqref{prob:pop_sdp} to its dual and passes to the solver MOSEK \cite{mosek}. As MOSEK is based on the primal-dual interior point method, thus it can generate a certificate for infeasibility \cite{andersen2000mosek}. We use the certificate returned by MOSEK to indicate the feasibility of the SDP. If \eqref{prob:pop_sdp} is infeasible, then its dual will be unbounded \cite{blekherman2012semidefinite, mosek}. Thus MOSEK will return large objective values even if it fails to generate the infeasibility certificate. MOSEK also returns \texttt{SLOW\_PROGRESS} flag if the problem does converge successfully, possibly around a minimum. As the SDP relaxation of POP does not ensure feasibility or local optimality if the SDP is not tight \cite{lasserre2015introduction}, for the drone landing case, we use the general purpose NLP solver IPOPT \cite{wachter2006implementation} for local search and the solution from SDP as the initial guess.
All experiments are launched on a desktop equipped with Intel i9-11900KF CPU and 128 GB memory. % \mgj{how much of this memory is necessary? 128 GB is probability overkill?} \tsl{40 step drone landing would use up all the memory if we only use CS. } \mgj{Readers will be curious about runtimes and this info and some discussions about them. When can we use this? Does it not seem to be appropriate for real-time applications or is it? Think about a reviewer who's very applied and wants to know about usage and results and will ignore the math. You want to have that reviewer on your side too.}

\subsection{Inverse kinematics for serial manipulator}
 % Feasibility detection for the PUMA 560 and comparison with the analytical IK solution. The green dots denote the pose is infeasible but falsely detected as feasible points. The black dots denote the feasible pose detected it feasible point. (a) Using \texttt{FEASIBLE\_POINT} certificate will falsely detect the infeasible point as feasible. (b) Using \texttt{INFEASIBILITY\_CERTIFICATE} only falsely detects feasible points as  \texttt{SLOW\_PROGRESS} according to the MOSEK solver. The black dots denote the pose is feasible, but the solver fails to detect but returns \texttt{UNKNOWN\_RESULT\_STATUS}.

% We consider the IK for pure kinematics problems and the rigid body trajectory optimization as the dynamics problem.
% \subsubsection{Inverse Kinematics}
% \subsubsection{Inverse kinematics problem for serial manipulator} $$X_k := \left[\begin{array}{cc} R_{k} & p_k\\  0   & 1 \end{array}\right] \in \mathrm{SE}(3)$$
We consider the inverse kinematics problem for $n$ Degrees of Freedom (DOF) serial manipulator with revolute joints. We use $X_k \in \mathrm{SE}(3)$ to represent the pose of $k$-th joint. 
Each joint is modeled as rotating about the local $z$-axis, i.e., $R_z$ described in \eqref{eq:joint_rotate} with the angle denoted by $c_k$ and $s_k$. Note that one can also use DH parameters with similar results. Thus, we have the kinematics chain:
\begin{equation}
    X_{k+1} = X_k A_{k+1} T_{k} , \quad X_0 = I,
\end{equation}
with 
\begin{equation}
    T_{k} := \left[\begin{array}{cc}
                   R_{k}^c & 0\\
                   0   & 1
                   \end{array}\right], 
   A_{k} := \left[\begin{array}{cc}
                   R^{z}_{k} & R^{z}_{k}  p^{c}_{k} \\
                   0   & 1
                   \end{array}\right]. 
\end{equation}
% In this formulation, joint $X_{k+1}$ and $A_k$ rotate about the $z$-axis of $X_k$ and then $T_k$ re-orientate the pose at the end of the arm. For joint angle constraints $\theta_{\min} \le \theta \le \theta_{\max}$ that is possibly asymmetric, we have the following \emph{linear} inequality
In this formulation, $p_k^c$ is the constant vector defining the arm, and $A_{k}$ is the action that describes the rotation of the arm about the $z$-axis of joint $X_k$. $T_k$ is a constant that re-orientates the pose for joint $X_{k+1}$. For joint angle constraints $\theta_{\min} \le \theta \le \theta_{\max}$, we have the following inequality:% \emph{linear}
\begin{equation}
\label{eq:joint_lim}
    \cos{(\theta-\frac{\theta_{\max} + \theta_{\min}}{2})} \ge \cos\frac{\theta_{\max} - \theta_{\min}}{2}.
\end{equation}
Then the IK problem for an $N$ DOF serial manipulator to reach the target pose $T_g \in \mathrm{SE}(3)$ is formulated as follows.

\begin{problem}[IK for N-DOF serial manipulator]
\label{prob:ik}
\begin{equation}
    \begin{aligned}
    \min_{\{c_k, s_k, X_k\}}& \quad  \sum_{k=1}^{N} (c_k - c_k^{r})^2 + (s_k - s_k^{r})^2 \\
    \text{s.t. } 
    &X_{k+1}= X_k A_{k+1} T_{k},\quad k = 0, 1, \dots, N-1. \\
    &X_k \in \mathrm{SE}(3),\quad c_{k+1}^2 + s_{k+1}^2 = 1, \\
    &c_{k+1}\bar{c}_{k+1}+s_{k+1}\bar{s}_{k+1} \ge c_{\lim}, \\ % M -\|p_{k+1}\|^2\ge 0,\\
        &X_0 = I, X_N = X_g.
    \end{aligned}
\end{equation}
\end{problem}

\begin{figure}[t]
    \centering
    \includegraphics[width=1\columnwidth]{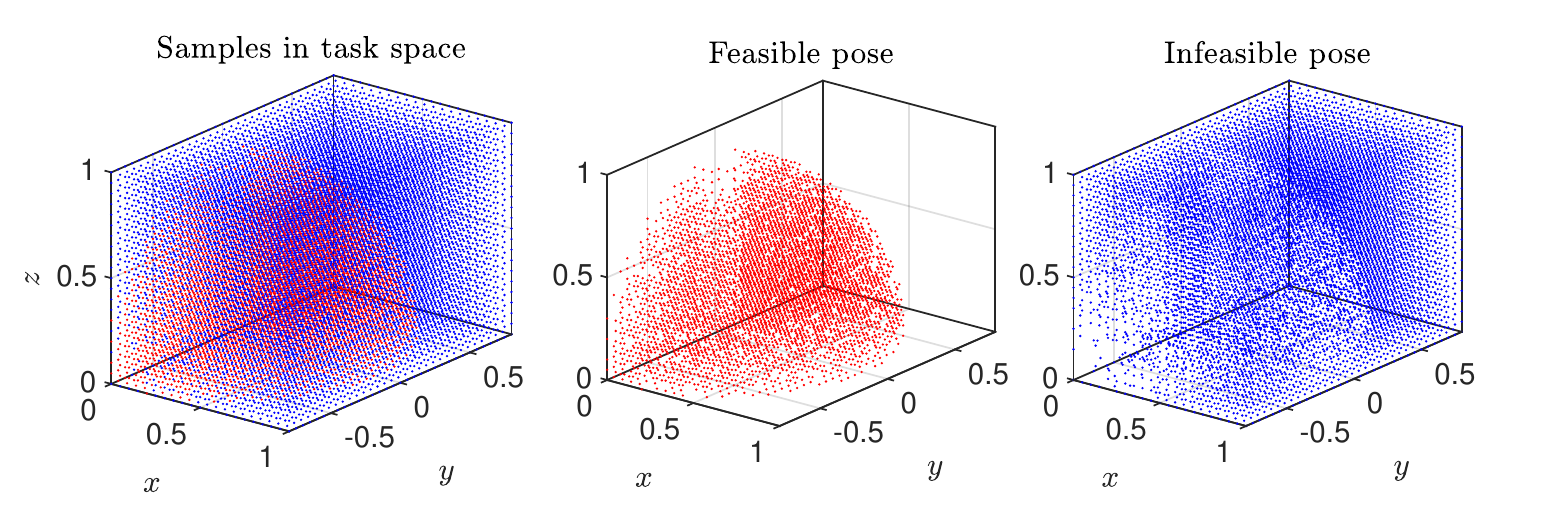}
    \caption{Samples in the workspace for 6-DOF PUMA 560 manipulator. We sampled $21\times 21\times 33 = 14553$ points with $5\ cm$ intervals in the workspace of size $1m\times 1m \times 1.6m$. The 4712 feasible poses are denoted as red, while the 9841 infeasible poses are denoted as blue.}% $4827$ $9726$
    \label{fig:puma_sample} 
\end{figure}
\begin{figure}[t]
    \centering
    \includegraphics[width=0.9\columnwidth]{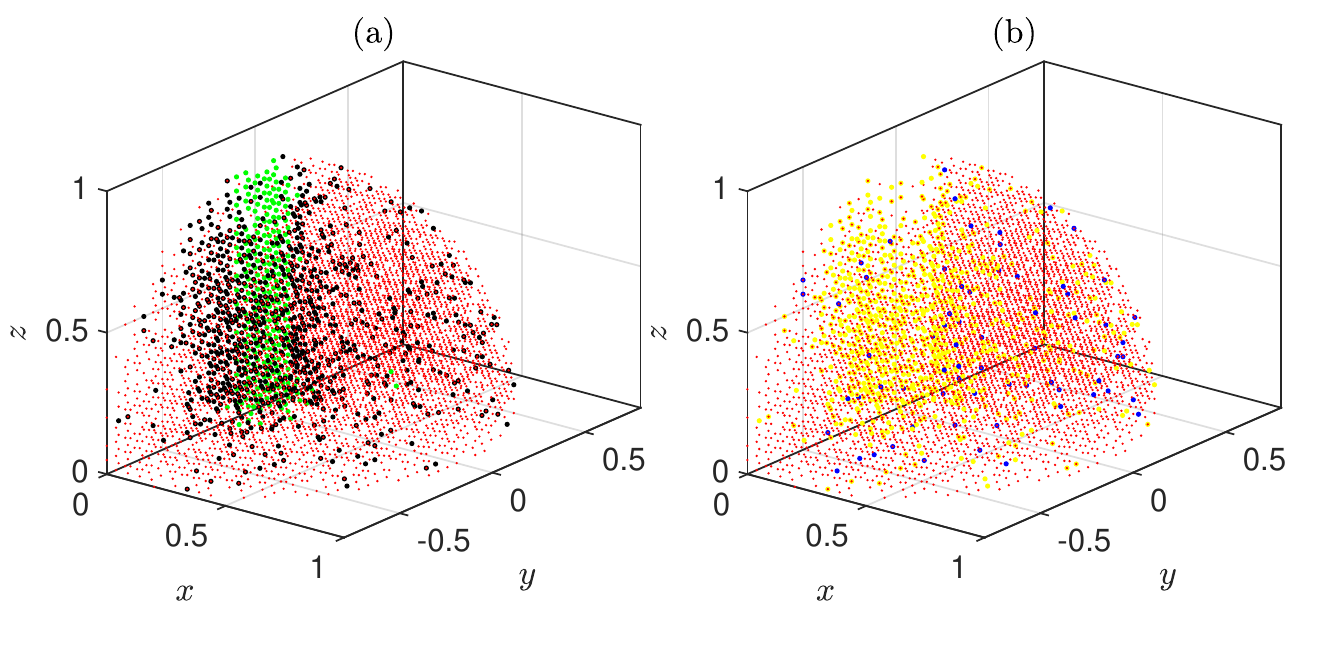}
    \caption{IK solution for 6-DOF PUMA 560 manipulator. (a) applied the CS at $\kappa=2$, the green dots denote the infeasible pose that the solver does not generate an infeasibility certificate, and the objective value is smaller than the threshold. The black dots denote the feasible poses that the solver does not converge and terminates as \texttt{SLOW\_PROGRESS}. (b) applied CS+MD at $\kappa=2$ with chordal extension. All the green dots in (a) can be certified as infeasible using this relaxation scheme via the certificate or unreasonably large dual problem values. The yellow dots denote the feasible poses with convergent SDP solution. The blue dots denote the case that is still not convergent at this relaxation scheme. When moving to $\kappa=3$, only one of the cases converges. } % Feasibility detection for the PUMA 560 and comparison with the analytical solution. The green dots denote the pose as infeasible but falsely detected as a feasible point. All these failed cases are terminated as \texttt{SLOW\_PROGRESS} according to the \texttt{MOSEK} solver.
    \label{fig:puma_feasibility} 
\end{figure}
Where $\bar{c}_k, \bar{s}_k$ and $c_{\lim}$ correspond to terms in the expansion of \eqref{eq:joint_lim}. The cost function is parameterized by a reference joint angle $c_k^r = 1$ and $s_k^r = 0$ in our case to ensure unique solutions. We tested the proposed algorithm on the 6-DOF manipulator PUMA 560 that has analytical IK solutions \cite{merat1987introduction}. As is shown in Fig.~\ref{fig:puma_sample}, we uniformly sample 14553 positions in the workspace with random rotations. We validate the algorithm based on the quality of infeasibility detection, and all the seven joint poses $X_k$ (including the base joint) directly recovered from SDP without refining. We consider the pose as infeasible if MOSEK returns an infeasibility certificate or unreasonably large objective value. We set the threshold as 1000, which is sufficiently larger than the maximal possible objective of Problem~\ref{prob:ik}. % As the maximal possible value of Problem \ref{prob:ik} is 24 for $N=6$, the SDP relaxation will only return values larger than 24 if it is infeasible. We also impose the norm constraints on position to make the feasible set more than compact \cite{lasserre2015introduction}.

The main results are summarized in Table \ref{table:puma_stat} and illustrated in Fig.~\ref{fig:puma_feasibility}. We first apply CS at $\kappa=2$. As shown in Fig.~\ref{fig:puma_feasibility}(a), we can see that 239 poses are infeasible but detected as feasible. We also have 879 feasible poses that the solver fails to converge and terminates as \texttt{SLOW\_PROGRESS}. Then we apply a slightly tighter relaxation, i.e., CS+MD at $\kappa = 2$, to improve the result. As is in Fig.~\ref{fig:puma_feasibility}(b), all the infeasible points can be certified, and 812 out of the 879 feasible cases converge. For the other 67 cases, we move to $\kappa=3$ with CS, but only 1 of them converges. Tighter relaxation, such as CS+MD at $\kappa=3$, will use up the memory. % , which suggests that the failure of these cases is more likely due to numerical issues.

\begin{table}[t]
\setlength\tabcolsep{4pt}
\centering
\scriptsize
\caption{The average performance of SDP relaxation on Puma 560 manipulator. We launch all the tests from a looser relaxation scheme and move to a tighter scheme for failed cases. The runtime is based on each individual relaxation scheme, while the other indices are accumulated from the lowest scheme. The pose error is based on cases successfully solved  by MOSEK.}
\begin{tabular}{l|ccc}
\hline
Relaxation order ($\kappa$) & 2 & 2 & 3 \\
Sparsity pattern & CS & CS+MD & CS \\
Average solution time (s) & $7.8$&  $918.7$& $804.8$\\
Maximum solution time (s) & $18.5$ & $2117.9$ &$1886.5$\\
Average joint orientation error (deg) & $6.93e^{-5}$ & $6.93e^{-5}$ & $6.93e^{-5}$\\
Maximal joint orientation error (deg) & $0.0231$ &$0.0231$ &$0.0231$\\
Average joint position error (cm) & $1.45e^{-5}$ & $1.45e^{-5}$ & $1.45e^{-5}$\\
Maximal joint position error (cm) & $0.0191$ & $0.0191$ & $0.0191$\\
Percentage of infeasibility detection & $97.57\%$  & $100.0\%$ & $100.0\%$ \\ % 97.73\%
Percentage of convergent SDP (feasible poses) & $81.35\%$ & $98.58\%$ & $98.60\%$ \\
Overall successful rate (all poses) & $92.32\%$ & $99.54\%$ & $99.55\%$ \\
% Convergent SDP (infeasible pose) & 92.32\% 100\%\\
\hline
% \vspace{-20pt}
\end{tabular}
 \label{table:puma_stat}
\end{table}
\begin{figure}[t]
    \centering
    \includegraphics[width=1\columnwidth]{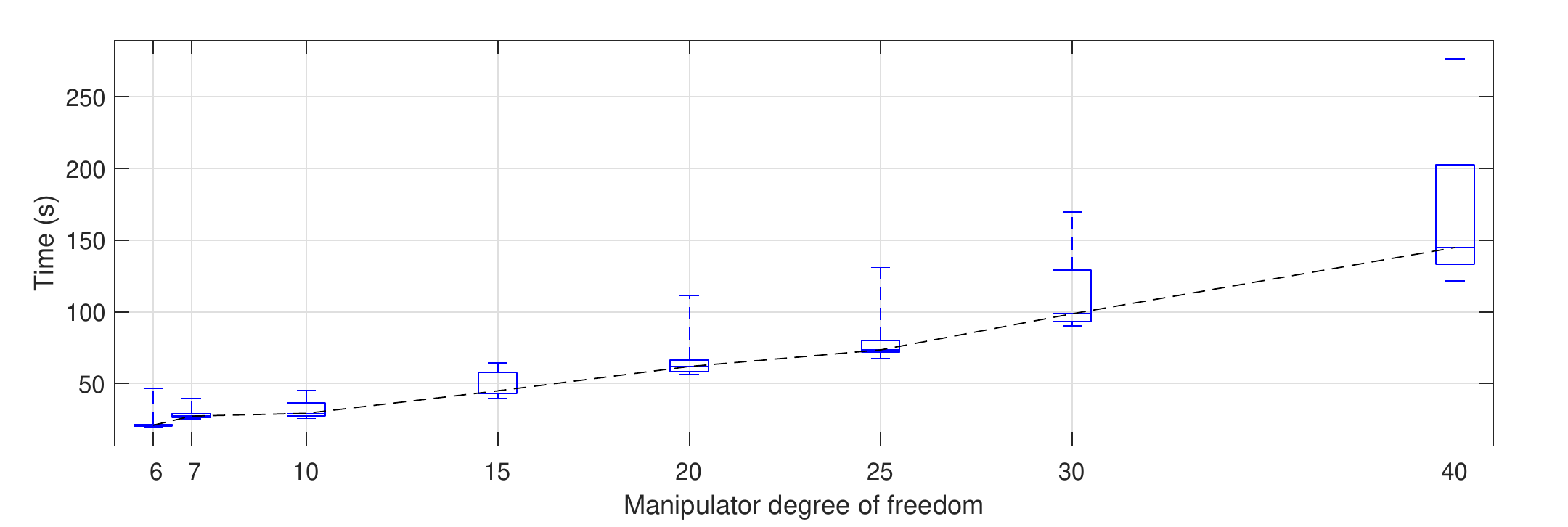}
    \caption{The runtime for the IK problem with DOF ranging from 6 to 40 tested on 50 randomly generated manipulators. We find that the time complexity is linear with respect to the DOF if CS is applied. The median value is denoted as connected by the black line. }
    \label{fig:IK_runtime}
\end{figure}

For the cases successfully solved by MOSEK, the average rotation and position errors are negligible. For the failed cases that terminated as $\texttt{SLOW\_PROGRESS}$, some cases have errors at a comparable level while some have position errors that can reach $1\ m$. Compared to the mix-integer programming-based method \cite{dai2019global}, the proposed algorithm has highly accurate solutions when the SDP converges to the optimum, as the kinematic model we use is exact. As \cite{dai2019global} can not avoid the inaccuracy due to the partition of $\mathrm{SO}(3)$ manifold, the position error can be centimeters. Both our method and \cite{trutman2022globally} are based on moment relaxation and suffer from similar numerical issues when solving the SDP, while the overall success rates are at the same level. % while 108 points are feasible, but the solver only returns \texttt{UNKNOWN\_RESULT\_STATUS}. 

Problem~\ref{prob:ik} has the same sparsity pattern as the kinodynamic planning Problem \ref{prob:kindyn_mp}. Thus, we further tested the runtime for Problem \ref{prob:ik} with DOF ranging from 6 to 40 with 50 randomly generated parameters $R^c_k$ and $p^c_k$ with CS at $\kappa = 2$ to evaluate the complexity of the algorithm. As is shown in Fig.~\ref{fig:IK_runtime}, the runtime grows linearly with respect to the DOF, which is consistent with the complexity analysis presented in \cite{lasserre2006convergent}. In comparison, \cite{dai2019global} has exponential computation time with respect to the planning horizon due to the combinatorial formulation. As \cite{trutman2022globally} is a dense formulation, the number of variables and degrees grows simultaneously. Considering the size of the moment matrix, \cite{trutman2022globally} has at least polynomial complexity with respect to the length of the kinematic chain. % As is shown in Table. 2 of paper \cite{dai2019global}, the average runtime to find a solution is exponential with respect to the binary variables when using mixed-integer programming. 
% The IK problem have the same structure as the kinodynamic planning \eqref{prob:kindyn_mp}. Thus, we further tested the runtime for manipulators with DOF ranging from 6 to 40 with 50 randomly generated parameters $R^c_k$ and $p^c_k$ with CS to evaluate the complexity of the algorithm.

\subsection{3D drone landing}

\begin{figure*}[t]
    \centering
    \includegraphics[width=2\columnwidth]{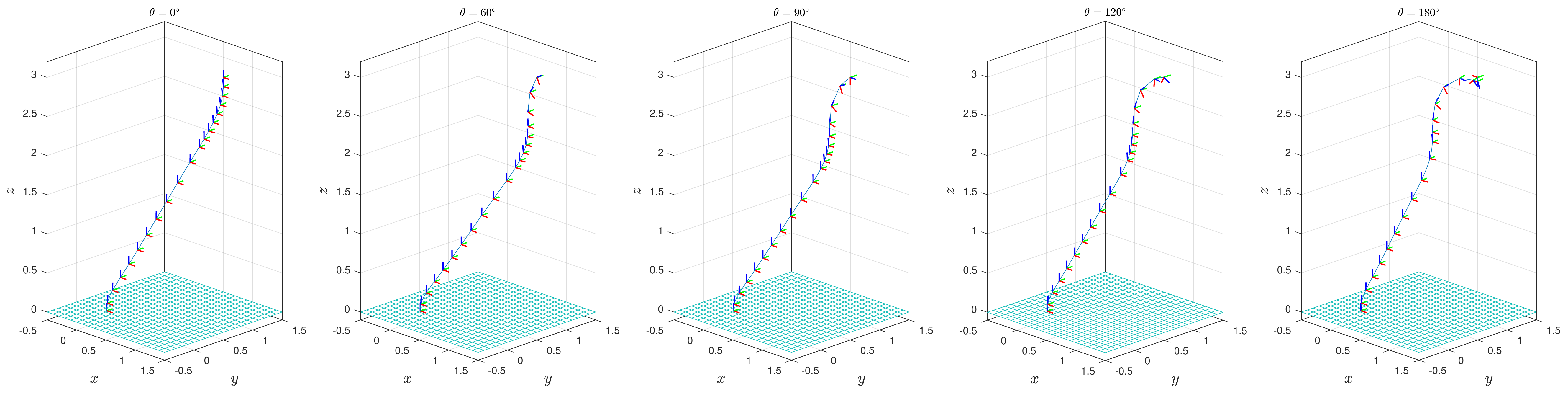}
    \caption{3D drone landing task in free space. The drone starts at an initial position $x=1, y=1, z=3$  with zero twists and different initial pitch angles. The robot is guided to land at the origin. The blue, red, and green axes indicate the $z$, $x$, and $y$ axes in the body frame. For a large initial pitch angle, the drone adjusts the orientation in the first few steps and then starts to move to the origin. Only part of the waypoints in the tail of trajectories are presented to avoid overlap. The statistics of these cases are presented in Table~\ref{table:drone_uncons}.}
    \label{fig:drone_traj_uncons}
    \vspace{-10pt}
\end{figure*}

\begin{figure*}[t]
    \centering
    \includegraphics[width=2\columnwidth]{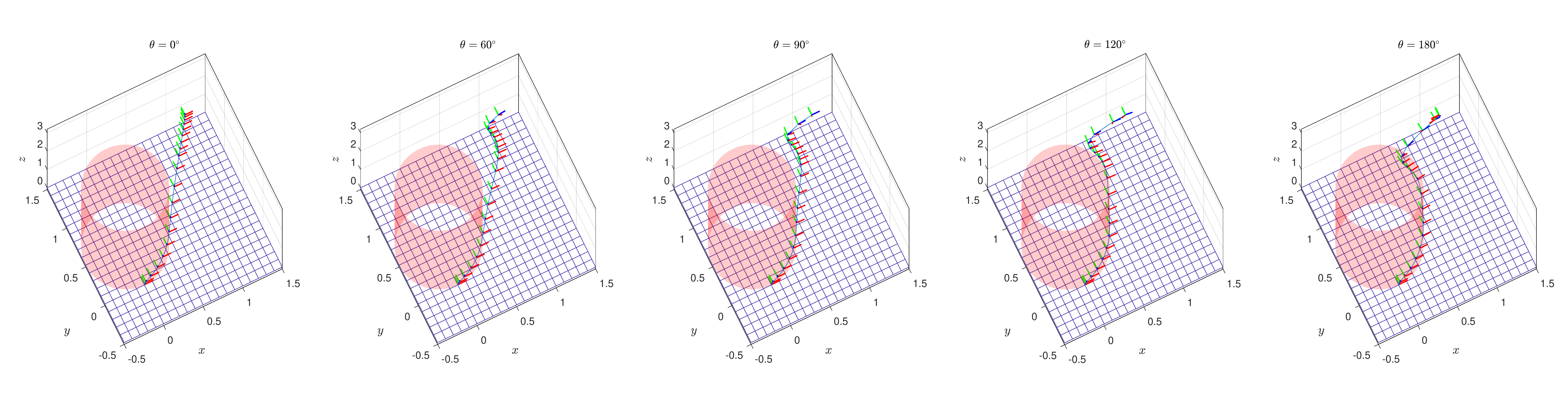}
    \caption{3D drone landing task with obstacle 1. Obstacle 1 blocks some of the waypoints computed from the free space task. All these plots converge to a locally feasible path after being refined by IPOPT, and the first 4 cases are certified as globally optimal solutions by metric $\epsilon$. The statistics of these cases are presented in Table~\ref{table:drone_obs1}.}
    \label{fig:drone_traj_obs1}
    \vspace{-10pt}
\end{figure*}

\begin{figure*}[t]
    \centering
    \includegraphics[width=2\columnwidth]{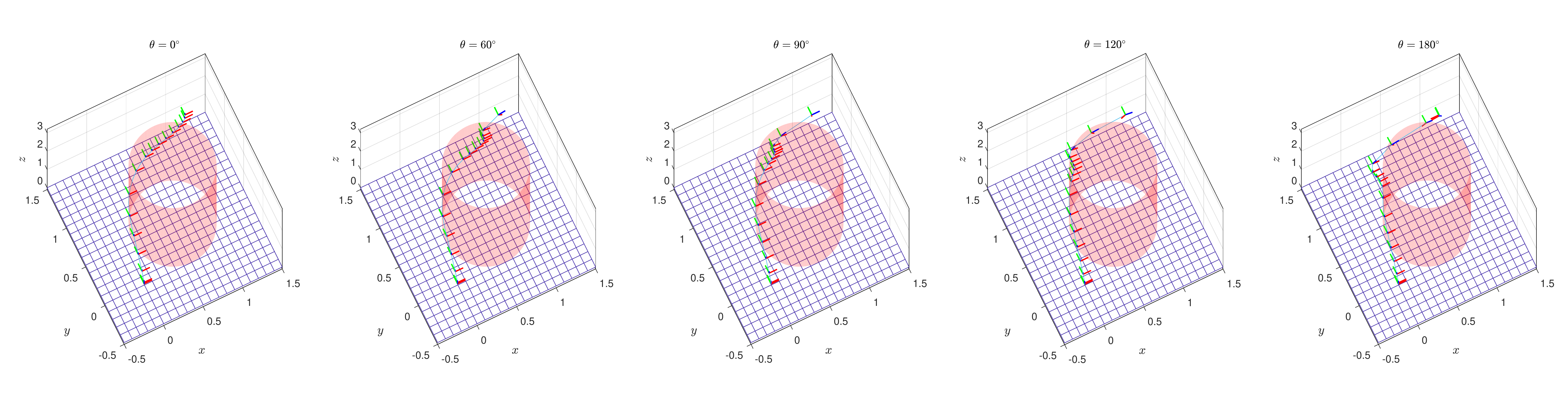}
    \caption{3D drone landing task with obstacle 2. Obstacle 2 blocks more waypoints in the free space task than obstacle 1, which makes the SDP harder to converge. All these cases converge to locally feasible solutions after being refined by IPOPT, and the first 4 cases have very small optimality gaps. The statistics of these cases are presented in Table~\ref{table:drone_obs2}.}
    \label{fig:drone_traj_obs2}
    \vspace{-10pt}
\end{figure*}

\begin{table}
\centering
\caption{Drone landing parameters}
\resizebox{\columnwidth}{!}{
\begin{tabular}{l|llc|l}
\cline{1-2}\cline{4-5}
Simulation constants     & Values                       &  & Control parameters     & Values.  \\ 
\cline{1-2}\cline{4-5}
Mass - $m$          & $0.5\ \ (\mathrm{kg})$                               &  & $P$      & $[100, 10, 100, 100]$            \\
Inertial - $I_b$    & $\mathrm{diag}(0.3, 0.2, 0.3)\ \ (\mathrm{kg / m^2})$&  & $Q$     & $[0.1, 10, 0.1, 1]$              \\
Gravity - $g$     & $[0,0,-9.81]^{\transpose}\ \ (\mathrm{m / s^2})$       &  & $U$     & $[0.1, 0.1]$            \\
Time step - $h$     & $0.1667s\  /\ 0.125s$      &  & $[\tau_{\min}, \tau_{\max}]$ &  $[-5, 5]$        \\
Planning horizon - $N$     & $30 \  /\ 40$      &  & $[f^z_{\min}, f^z_{\max}]$ &  $[-\infty, \infty]$        \\
Obstacle 1   & $x^2 + (y-0.5)^2 \ge 0.25$  && $p_0, v_0, F_0$ &  $[1,1,3]^{\transpose}, 0_{3\times 1}, I_3$        \\
Obstacle 2   & $(x-0.6)^2 + (y-0.5)^2 \ge 0.25$&  & $R_0 = R_y(\theta)$ & $\theta=0^{\circ}, 60^{\circ}, 90^{\circ}, 120^{\circ}, 180^{\circ}$        \\
Height constraints  & $z \ge 0$&  &  &        \\
\cline{1-2}\cline{4-5}
\end{tabular}}
\label{table:drone_param}
\end{table}

\begin{table*}%[b]
\setlength\tabcolsep{4pt}
\centering
% \small
% \tiny
\scriptsize
\caption{Statistics of drone landing in free space. TS+block can solve all of the cases with pitch angles smaller than $120^\circ$. For the harder cases, CS is needed to induce tighter relaxation. These cases are illustrated in Fig. \ref{fig:drone_traj_uncons}. }
\begin{tabular}{c|ccc|ccc|ccc|ccc|ccc}
\hline
$R_{y,0}$&  & $0^{\circ}$ &  &  &$60^{\circ}$  &  &  & $90^{\circ}$ &  &  & $120^{\circ}$ &  &  &$180^{\circ}$  \\ \hline
 Sparsity  & TS+MD  & TS+block  & CS & TS+MD  & TS+block  & CS& TS+MD  & TS+block  & CS& TS+MD  & TS+block  & CS & TS+MD  & TS+block & CS \\
 Suboptimality $\epsilon$ & $0.0101$  &$\le 1e^{-4}$   &N/A  &$0.0235$  &$\le 1e^{-4}$  &N/A  &$0.0589$  &$\le 1e^{-4}$       &N/A  &$0.3767$  &$0.0046$  &$\le1e^{-4}$  & $0.4102$ &$0.2054$  &$0.0136$\\
Rank condition $\delta$   & $0.0968$  &$\le 1e^{-4}$   &N/A  &$0.1702$  &$\le 1e^{-4}$  &N/A  &$0.4918$  &$6.302e^{-4}$      &N/A  &$0.8301$&$0.5824$    &$\le1e^{-4}$  &$0.9049$  &$0.9653$  &$0.6504$\\ 
 Runtime (s)    & $397.1$   &$2299.6$        &N/A  &$431.4$   &$2185.9$       &N/A  &$381.3$   &$2298.5$        &N/A  &$373.6$& $2868.3$    &$6662.0$  &$346.7$   &$3646.5$  &$6820.4$\\ \hline
\end{tabular}
 \label{table:drone_uncons}

\end{table*}

\begin{table*}%[b]
\setlength\tabcolsep{4pt}
\centering
% \small
% \tiny
\scriptsize
\caption{Statistics of drone landing with obstacle 1. All cases can be tightly solved except the last one. These cases are illustrated in Fig. \ref{fig:drone_traj_obs1}.}
\begin{tabular}{c|ccc|ccc|ccc|ccc|ccc}
\hline
$R_{y,0}$&  & $0^{\circ}$ &  &  &$60^{\circ}$  &  &  & $90^{\circ}$ &  &  & $120^{\circ}$ &  &  &$180^{\circ}$  \\ \hline
 Sparsity   & TS+MD         & TS+block         & CS   & TS+MD        & TS+block         & CS  & TS+MD       & TS+block         & CS  & TS+MD       &TS+block     & CS      & TS+MD       & TS+block    & CS \\
Suboptimality  $\epsilon$ & $0.0130$   &$\le 1e^{-4}$  & N/A  & $0.0252$  &$\le 1e^{-4}$  &N/A  &$0.0608$  &$\le 1e^{-4}$  &N/A  &$0.3772$  &$0.0070$  &$\le1e^{-4}$  &$0.3810$  &$0.2083$  &$0.0205$\\
 Rank condition $\delta$   &$0.0965$    &$\le 1e^{-4}$  &N/A   &$0.1696$   &$\le 1e^{-4}$  &N/A  &$0.4805$  &$4.631e^{-4}$  &N/A  &$0.8240$  &$0.5687$  &$\le1e^{-4}$  &$0.9194$  &$0.9661$  &$0.7051$\\
 Runtime (s)    & $424.0$    &$2598.5$       &N/A   &$361.6$    &$2549.9$       &N/A  &$426.7$   &$3246.1$       &N/A  &$433.5$   &$2863.2$  &$7039.4$  &$409.2$  &$3834.5$   &$8641.7$\\ \hline
\end{tabular}
 \label{table:drone_obs1}
\end{table*}

\begin{table*}%[b]
\setlength\tabcolsep{2.5pt}
\centering
% \small
% \tiny
\scriptsize
\caption{Statistics of drone landing with obstacle 2. Due to the configurations of the obstacle, this task is more challenging, but the overall optimality gap using CS is small except for the $180^\circ$ case. These cases are illustrated in Fig. \ref{fig:drone_traj_obs2}. }
\begin{tabular}{c|ccc|ccc|ccc|ccc|ccc}
\hline
$R_{y,0}$&  & $0^{\circ}$ &  &  &$60^{\circ}$  &  &  & $90^{\circ}$ &  &  & $120^{\circ}$ &  &  &$180^{\circ}$  \\ \hline
 Sparsity   & TS+MD             & TS+block    & CS     & TS+MD       & TS+block    & CS  & TS+MD       & TS+block    & CS & TS+MD  & TS+block  & CS & TS+MD  & TS+block & CS \\
 Suboptimality $\epsilon$ &$0.0499$        &$0.0275$  &$7.731e^{-4}$       &$0.0580$  &$0.0307$  &$0.0035$     &$0.0802$  &$0.0473$  &$\le 1e^{-4}$  &$0.3846$  &$0.0154$  &$\le 1e^{-4}$  &$0.4231$  &$0.2127$  &$0.0167$\\
  Rank condition $\delta$  &$0.1013$        &$0.0738$  &$0.0369$        &$0.1776$  &$0.0661$  &$0.1207$     &$0.5019$  &$0.1849$  &$0.0016$  &$0.8348$  &$0.6049$  &$\le 1e^{-4}$  &$0.9088$  &$0.9632$  &$0.6712$\\
  Runtime (s)   &$391.1$     &$3295.4$  & $8314.3$       &$360.5$  &$3111.7$  &$9310.6$     &$424.2$   &$3471.0$   &$7241.5$  &$416.2$   &$3350.2$  &$7700.8$  &$429.1$   &$3686.7$  &$8180.6$\\ \hline
\end{tabular}
  \label{table:drone_obs2}
\end{table*}

\begin{table*}%[b]
\setlength\tabcolsep{2.5pt}
\centering
% \small
% \tiny
\scriptsize
\caption{Statistics of drone landing task solved by IPOPT. We perturb the best optimality guaranteed SDP solution with different noise levels to obtain the initial guess. The average suboptimality is evaluated on each task with 5 initial pitch angles. For each test case, we sample the initial error 10 times. Thus the average suboptimality is evaluated by feasible solutions among the 50 random samples. We show that with a small noise level, IPOPT converges to the best solution obtained by SDP. However, as the noise level increases, the optimality gap grows significantly, and even local feasibility is not guaranteed. We set the computational budget as 1e4 iterations.}
\begin{tabular}{c|ccc|ccc|ccc|ccc|ccc}
\hline
Initialization noise $\Delta$ &  & 0.001 &  &  & 0.01  &  &  & 0.1 &  &  & 0.5 &  &  & 1  \\ \hline
 Task                             & Free        & Obs-1    & Obs-2              & Free     & Obs-1    & Obs-2& Free    & Obs-1    & Obs-2  & Free     & Obs-1    & Obs-2  & Free      & Obs-1    & Obs-2 \\
 Average Suboptimality $\epsilon$ &$\le1e^{-4}$ &$\le1e^{-4}$  &$\le1e^{-4}$  &$\le1e^{-4}$  &$\le1e^{-4}$  &$\le1e^{-4}$ &$0.0115$  &$0.0134$  &$\le1e^{-4}$  &$0.3644$  &$0.4757$  &$0.2406$  &$0.7727$  &$0.7543$  &$0.5652$\\
 Timeout (out of 50)  &0 &0  &0  & 0 &0  &0 &0 &0 &0  &12  &8  &2  &25  &15  &6\\
 Infeasibility (out of 50)  &0 &0  &0  & 0 &0  &0 &0 &0 &0  &13  &11  &13  &16  &23  &13\\
\hline
\end{tabular}
\label{table:ipopt_test}
\end{table*}

In this task, we consider landing a drone from an initial pose while avoiding obstacles. We assume that the drone is controlled by torque $\tau \in \mathbb{R}^3$ and total thrust force $f^z \in \mathbb{R}$ along the $z$-axis, both in the body frame. Thus we have the kinematics model \eqref{eq:lgvi_kinematic} and the forced drone dynamics:
\begin{equation}
\begin{aligned}
    & I^bF_{k+1}^{\transpose} - F_{k+1}I^b=F_k^{\transpose}I^b - I^bF_{k} + h^2\tau^{\times}_{k+1}, \\
    &mv_{k+1} = mF^{\intercal}_kv_{k} + h(Bf^z_{k+1} + mR_{k+1}g),
\end{aligned}
\end{equation}
with $ B = [0,0,1]^{\transpose}$. We then define the quadratic cost:
\begin{equation}
    \begin{aligned}
        \Phi(Y_N) = &P_1 \|R_N - I\|^2_{F,I} + P_2 \|F_N - I \|^2_{F, I} + \\
        &P_3\|p_N\|^2 + P_4\|v_N\|^2,
    \end{aligned}
\end{equation}
\begin{equation}
    \begin{aligned}
        L(&Y_k, u_{k+1}) = Q_1 \|R_k - I\|^2_{F,I} + Q_2 \|F_k - I \|^2_{F, I} + \\
        &Q_3\|p_k\|^2 + Q_4\|v_k\|^2 + U_1 \|\tau_{k+1}\|^2 + U_2 \|f^{z}_{k+1}\|^2,
    \end{aligned}
\end{equation}
where $\|X\|_{F,P} = \sqrt{\operatorname{tr}(X^{\transpose}PX)}, P\ge 0,$ is the weighted Frobenius norm. The system parameters and cost functions are shown in Table. \ref{table:drone_param}. We set the drone at an initial position with different pitch angles. We launch the trajectory optimization with one of the obstacles or in free space. Obstacle 2 blocks more waypoints of the trajectory planning in the free space compared to obstacle 1. The whole planning horizon is 5 seconds with 40 steps. For the CS cases without TS, the SDP has too many variables that used up the memory, so we reduced the step number to 30. We start with TS+MD and continue to TS+block or CS if $\epsilon$ is large for a former relaxation scheme. Empirically, we find that $\epsilon \le 1e^{-3}$ results in tight moment relaxation that also guarantees the feasibility of the solution. Thus, we set $\epsilon \le 1e^{-3}$ as a threshold to decide whether to continue to a tighter relaxation scheme.

The planned trajectories of all cases after being refined by IPOPT are illustrated in Fig.~\ref{fig:drone_traj_uncons}, \ref{fig:drone_traj_obs1} and \ref{fig:drone_traj_obs2}. The statistics of the planned trajectories are presented in Table \ref{table:drone_uncons}, \ref{table:drone_obs1}, and \ref{table:drone_obs2}.  For all tests, TS+MD does not provide an optimality gap smaller than $0.01$. For simple cases in free space or with obstacle 1, TS+block can solve all the cases with $\theta \le 90^{\circ}$. For $\theta = 120^{\circ}$, moving to CS also generates certifiable optimality values. While for $\theta = 180^{\circ}$, the CS does not provide certifiable optimal values. For hard cases with obstacle 2, more numerical issues are presented. Even the CS sparsity pattern does not provide certifiable solutions for $\theta = 60^{\circ}$ and $\theta = 180^{\circ}$. 

As the solver returned \texttt{SLOW\_PROGRESS} for these failed cases, we do not know if the SDP relaxation itself is tight at this order for these initial conditions. However, the solution is still a good initial guess for local search by IPOPT. We can also find that the magnitude of $\epsilon$ and $\delta$ are closely correlated to each other. When $\epsilon\le 1e^{-4}$, the rank-one condition will likely satisfy as $\delta$ is also small.

% The planned trajectories of all the cases after refined by IPOPT are illustrated in Fig.~\ref{fig:drone_traj_uncons}, \ref{fig:drone_traj_obs1} and \ref{fig:drone_traj_obs2}. The statistics of the planned trajectories are presented in Table \ref{table:drone_uncons}, \ref{table:drone_obs1}, and \ref{table:drone_obs2}.  For all the tests, the TS+MD does not provide an optimality gap smaller than $0.01$. The TS+block can solve simple cases such as $\theta \le 90^{\circ}$ for free space and obstacle 1. For hard cases, $\theta = 120^{\circ}$, CS can solve most of them with the required accuracy except for $\theta = 180^{\circ}$. As the solver returned \texttt{SLOW\_PROGRESS} for these failed cases, we do not know if the SDP relaxation itself is tight at this order for these initial conditions. However, the solution is still a good initial guess for local search by IPOPT. We can also find that the magnitude of $\epsilon$ and $\delta$ are closely correlated to each other. When $\epsilon\le 1e^{-4}$, the rank-one condition is likely to satisfy as the $\delta$ is also small.

We further compare the proposed methods with local solvers. We compute POP using IPOPT and initializations with different qualities. As the moment relaxation already gives an initial guess with guaranteed lower bound by suboptimality metric $\epsilon$, we perturb the refined solution with the smallest $\epsilon$ as our initial guess. For the orientation $R_k$ and pose change $F_k$, we perturb it by rotating with a random angle as
$\tilde{R}_k =  R_k\exp{(\zeta)}$
where $\zeta\sim N(0_{3,1}, \Delta^2 I_3)$ is white Gaussian noise. We perform element-wise random perturbation for other variables, such as $\tilde{v}_k = v_k + \zeta$. Then we assign different $\Delta$ to represent the quality of the initial guess. We mainly evaluate the IPOPT solution via the optimality gap $\epsilon$ compared to the best-known refined solutions by SDP and the convergent status of IPOPT. For each $\Delta$, we sample 10 times for all the 5 initial conditions and tasks. The result is presented in Table~\ref{table:ipopt_test}. The local solver highly relies on the initial guess quality, as shown in Table~\ref{table:ipopt_test}. However, such a good initial guess is not always available in practice. Though the local search method is fast, it is possible that the solver converges to infeasible points.%, which is more deteriorate. % For example, straight line initialization \cite{ratliff2009chomp} is usually used. Thus one should not expect global optimality when using local search methods. 

\section{Discussions}
We have shown that Lasserre's hierarchy at the second order can provide rank-one globally optimal solutions for most testing cases of the two case studies. More than 99.5\% of the IK problems can be solved, while the remaining cases are subject to numerical issues. By LGVI, the drone dynamics can be formulated as exact quadratic polynomials, which is not possible using a conventional integration scheme or approximations. The SDP converges well for easy cases with certified globally optimal solutions while suffering from numerical issues for hard cases. From the point of view of computational complexity, the worst-case for arbitrary obstacle configuration still results in SDPs of computationally intractable size. Thus, a convex representation of free space \cite{amice2023finding}, obstacles \cite{han2022non, jasour2019risk, thirugnanam2022safety}, or risk-bounded contours of \cite{jasour2013convex, jasour2016convex, jasour2015semidefinite} should be explored. % {\color{brown}In  moment-based SDP has been applied to address the risk-bounded control of stochastic polynomial systems.}
 % To solve the numerical issues, SDP solvers with high accuracy for the rank one solution, such as \cite{yang2022inexact}, should be applied in the future. % As is presented in \cite{yang2022inexact}, blending the fast local search by nonlinear solvers and global convergence from moment relaxation is also a promising technique. 

A more intriguing future direction is to find conditions when such Lie group-based motion planning problems can be tightly solved via Lasserre's hierarchy at the second order, despite the numerical challenges. Such efforts would be essential for convex kinodynamic motion planning using the full dynamics model. The LGVI-based formulation can be extended to multi-body systems when additional constraints and multipliers are added \cite{leyendecker2008variational}. Nonholonomic systems can also be modeled by LGVI, while the discretization of nonholonomic constraints needs more attention \cite{fernandez2012variational, kobilarov2010geometric}. More generally, LGVI is aligned with symmetry-preserving methods in robotics~\cite{ghaffari2022progress} and its applications in various robot perception and control problems from a promising research direction. 

The memory and time cost of the proposed algorithm is still high for real-time applications. In view of global optimization, the complexity is linear w.r.t the planning horizon and polynomial order w.r.t the system dimension (consider the size of the moment matrix) when the relaxation order is determined. Such property is an improvement compared to other global optimization methods, such as combinatorial optimization with exponential complexity. To improve the scalability for real-time deployment, combining the global convergence property of SDP \cite{yang2022inexact} and fast local search on Lie groups \cite{brockett1991dynamical,bloch1992completely,clark2021nonparametric,teng2022lie,teng2022error}, should be considered in the future. 

% faster solver, such as \cite{yang2022inexact}, which boosts the SDP via the fast local search by nonlinear solvers, should be applied in the future. \tsl{Incorporated some reply from rebuttals.}
% Though the runtime of the proposed algorithm is too long, such large-scale motion planning problems is impossible to be formulated as SDP without the proposed formulation. 

% Though purely using the CS sparsity pattern in the IK task can not generate an infeasibility certificate or fail to converge to the optimum for all the cases, a slightly tighter relaxation at order two can solve most of the remaining failed cases. By LGVI, the drone dynamics can be formulated as exact quadratic polynomials, which is not possible using a conventional discretization scheme or approximations. The SDP converges well for the easy cases with certified globally optimal solutions, while terminated with \texttt{SLOW\_PROGRESS} for harder cases. However, these nonconvergent cases still provide a good initial guess that enables fast local search with small optimality gaps. 

 % As the CS pattern guarantees the same convergent property as dense formulation \cite{lasserre2006convergent}, the problem is more likely to be the ill-posed SDP caused by the poses. All the tasks (except one) with an initial pitch angle smaller than 120 degrees can be solved tightly with an optimality gap at the level of $1e^{-4}$. For the task with an initial pitch angle of 180 degrees, the solver did not provide suboptimality smaller than 0.01.

% numerical issues. 
% multi slution for Ik
% non-trivial lower bound
\section{Conclusions}
In this paper, we present the novel result: by formulating the robot dynamics model on Lie groups, one can convert the motion planning problem to polynomial optimization problems that can be solved via Lasserre's hierarchy. We show that the proposed formulation converts rigid body dynamics as exact quadratic polynomials. We further formulate the motion planning problem as a sparse moment relaxation problem. Attributed to the low-order and sparse formulation, the resulting SDP has linear complexity with respect to the planning horizon and is computationally tractable for the current solvers. The case study on the inverse kinematics for serial manipulators and 3D drone landing problems suggests that the proposed formulation can successfully provide certified globally optimal solutions for most cases. % \tsl{Softened the tightness claim here.}% As the Lie group formulation is general for rigid body motions, such formulation is generalizable for more complicated robots. 
% \clearpage

% \appendix
\begin{appendices}

\section{POP as infinite-dimensional linear programming}
\label{appx:inf-lp}
Problem \ref{prob:pop} can be converted to the following infinite dimensional linear programming problem over the space of measure~\cite{lasserre2001global, lasserre2015introduction}:
\begin{problem}[Infinite dimension linear programming]
\begin{equation}
\label{prob:inf-lp}
\begin{aligned}
    p^*&:=\inf_{\mu\in\mathcal{M}^+(\mathbb{K})} \int_{\mathbb{K}} p(x) d\mu, 
\end{aligned}\tag{LP}
\end{equation}
with $\mathbb{K}$ the feasible set defined in \eqref{prob:pop}, $\mathcal{M}(\mathbb{K})$ the set of vector space of finite signed Borel measure and $\mathcal{M}^+(\mathbb{K})$ the convex cone of nonnegative finite Borel measure on $\mathbb{K}$. 
\end{problem}
Then, the optimization problem \eqref{prob:pop} is equivalent to finding the Dirac measure $\delta_{x^*}$ that is associated with the minimizer $x^*$ of \eqref{prob:pop}. Recall that the Dirac measure has the property $\int_{\mathbb{K}} p(x) d \delta_{\bar{x}} = p(\bar{x}), $
that enables one to select the value of $p(x)$ at a given point $\bar{x}$.

% \end{appendices}
% \begin{appendices}

\section{Sparse moment relaxation of POP}
\label{appx:sparse-pop}
Though Lasserre's Hierarchy enables one to approximate \eqref{prob:pop} by \eqref{prob:pop_sdp}, the size of \eqref{prob:pop_sdp} increases dramatically as $\kappa$ and $n$ increase. Therefore, one should fully explore the structure of the problem to reduce the computational burden. For many applications in control and planning satisfying the Markov assumption, only states at consecutive time steps appear in the system dynamics. The cost function is usually the sum of stage costs that only contain states within one step. Motivated by this observation, we introduce the correlative sparsity. % that appears in these problems.  % This structure could help use to search over a group of smaller moment matrices that contain the related variables instead of a large dense moment matrix for all the states. A $\kappa$-order dense moment matrix is formed by $s(2\kappa)$-dimensional moment sequence, where the variable grows exponentially with respect to $\kappa$ and at polynomial order with respect to $n$. 

We define the index set $I_0 = \{1, \dots, n\}$ be the union of $\cup_{k=1}^q I_k$ of $q$ subsets $I_k\subset I_0$ that partition the variable $x$. For arbitrary $I_k\subseteq I_0$, let $\mathbb{R}[x(I_k)]$ denote the ring of polynomials in the variable $x(I_k) \in \{x_i|i\in I_k\}$. We also define the index set $J=\{1,\dots,m\}$ that is partitioned in to $q$ different disjoint sets $J_k, k = 1,\dots,q$ to group the constraints $g_j, j=1,\dots,m$. 
%Then, we assume \eqref{prob:pop} have the following property:
\begin{assumption}[Sparse structure of \eqref{prob:pop}, \cite{lasserre2006convergent, lasserre2015introduction}]
\begin{enumerate}
    \item[]
    \item For feasible set $\mathbb{K}$, there is a large number $M$, such that $\|x\|_{\infty}\le M$ for $\forall x \in \mathbb{K}$. 
    \item For every $j \in J_k$, $g_j \in \mathbb{R}[x(I_k)]$, such that each constraint $g_j(x) \ge 0$ only involves variables in the set $x(I_k) = \{x_i | i \in I_k\}$. 
    \item The objective function $p(x) \in \mathbb{R}[x]$ can be written as $p(x) = \sum_{k=1}^{q} f_k$, with $f_k \in \mathbb{R}[x(I_k)], k = 1,\dots,q. $
    \item The index set $I_k$ satisfy the running intersection property: 
    $$\forall k=1,\dots, q-1, \exists s \le k, I_{k+1} \cap \left(\cup_{j=1}^kI_j\right) \subseteq I_s. $$ 
\end{enumerate}
\label{assumption:sparse_structure}
\end{assumption}

If \eqref{prob:pop} satisfies the assumptions, the following sparse moment relaxations can dramatically reduce the problem size. 
\begin{problem}[Sparse moment relaxation \cite{lasserre2006convergent, lasserre2015introduction}]
\begin{equation}
\begin{aligned}
\label{prob:pop_sparse_sdp}
    \rho_{\kappa}^*&:=\inf_{y\in \mathbb{R}^{s(2\kappa)}} \mathcal{L}_y(p) \\
    s.t. \quad &M_{\kappa}(y, I_k) \ge 0, \\
    \quad &M_{\kappa-d_i}(g_jy, I_k) \ge 0, \\
    &j \in J_k, k = 1, \dots, q.
\end{aligned}\tag{sparse-SDP}
\end{equation}
\end{problem}
% Every constraints $g_i$ in $\mathbb{K}$ contain variables $\{X_i | i \in I_k\}$. The objective function can be written as $p(x) = p_1 + p_2 + \dots + p_m$, where each $p_k$ only uses variables $\{X_i | i \in I_k\}$.
Where $M_{\kappa}(y,I_{k})$ denotes the moment matrix formed by the variables that appear in the set $I_k$. We also have a slightly different rank condition and the special rank-one case for the sparse moment relaxation. 
\begin{theorem}[Rank condition for \eqref{prob:pop_sparse_sdp} \cite{lasserre2006convergent, lasserre2015introduction}]
    \eqref{prob:pop_sparse_sdp} is tight, if:
\begin{enumerate}
    \item If Assumption \ref{assumption:sparse_structure} is satisfied for \eqref{prob:pop}, and, 
    \item $\operatorname{rank}(M_{\kappa}(y_{\kappa}^{*}, I_k)) = \operatorname{rank}(M_{{\kappa} - d_g}(y_{\kappa}^{*}, I_k))$, \mbox{$k=1,\dots,p$}, and, 
    \item $\operatorname{rank}(M_{\kappa}(y_{\kappa}^{*}, I_{jk})) = 1$ for all pairs $(i, k)$ with \mbox{$I_{jk}:=I_j\cap I_k \neq \emptyset $}.
\end{enumerate}
\label{theorem:sparse_rank}
\end{theorem}
\begin{remark}[Rank one optimality condition for \eqref{prob:pop_sparse_sdp}]
\label{remark:sparse_rank_one}
    For the special case of rank one condition, we will only need to check the first two conditions in Theorem \ref{theorem:sparse_rank} as any $M_{\kappa}(y,I_{jk})$ can become a principle submatrix of the $M_{\kappa}(y,I_{k})$ or $M_{\kappa}(y,I_{j})$ after proper invertible row and column permutations. Thus, $M_{\kappa}(y,I_{jk})$ is rank one if $M_{\kappa}(y,I_{k})$ or $M_{\kappa}(y,I_{j})$ are rank one matrix. 
\end{remark}

\end{appendices}

% \clearpage
{
\small 
\bibliographystyle{unsrtnat}
\bibliography{bib/strings-full,bib/ieee-full,bib/references}
}
% \bibliography{references}
%

\end{document}

%% file: definitions.tex
\usepackage{graphicx}
\usepackage{caption}
\graphicspath{
    {figures/}
}

\usepackage{amsmath,amssymb,enumerate}

\usepackage{amsthm}
\usepackage{setspace}
\usepackage{booktabs}
\usepackage[usenames,dvipsnames,svgnames,table]{xcolor}
\usepackage{mathtools}
\usepackage{algorithm, algorithmicx, algpseudocode}
\usepackage{blindtext}
\usepackage{gensymb}
\usepackage{xparse}
\usepackage{lipsum}
\usepackage{mathrsfs}
\usepackage[mathscr]{euscript}
\usepackage{times}
\usepackage[numbers,sort&compress]{natbib}
\usepackage{multicol}
\definecolor{darkgreen}{rgb}{0,0.6,0}
\usepackage[bookmarks=true,colorlinks=true,pdfpagemode=UseNone,citecolor=darkgreen,linkcolor=black,urlcolor=BrickRed,pagebackref]{hyperref}
\usepackage[caption=false,font=footnotesize]{subfig}
\usepackage{amsfonts}
\usepackage{cleveref}
\usepackage[utf8]{inputenc}
\usepackage[T1]{fontenc}
\usepackage{textcomp}
\usepackage{arydshln}
\usepackage{tabu}
\usepackage{cuted}
\usepackage{flushend}

\newtheorem{problem}{Problem}
\newtheorem{theorem}{Theorem}

\newtheorem{remark}{Remark}

\newtheorem{assumption}{Assumption}

\renewcommand{\thefigure}{\arabic{figure}}

\definecolor{note}{rgb}{0.1,0.1,1}
\definecolor{rephase}{rgb}{0.15,0.7,0.15}
\definecolor{bag}{rgb}{0.6,0.6,0.2}

\makeatletter
\renewcommand*\env@matrix[1][c]{\hskip -\arraycolsep
  \let\@ifnextchar\new@ifnextchar
  \array{*\c@MaxMatrixCols #1}}
\makeatother

\newcommand{\transpose}{\mathsf{T}}

\makeatletter
\newcommand{\mathleft}{\@fleqntrue\@mathmargin0pt}
\newcommand{\mathcenter}{\@fleqnfalse}
\makeatother